\documentclass[twocolumn, switch]{article} 
\usepackage{preprint}
\usepackage{amsmath,amssymb,amsfonts}
\usepackage{algpseudocode}
\usepackage{graphicx}
\usepackage{comment}
\usepackage{url} 
\usepackage{textcomp}
\usepackage{listings}
\usepackage{algorithm}
\usepackage{url}            
\usepackage{booktabs}       
\usepackage{nicefrac}       
\usepackage{microtype}      
\usepackage{lipsum}
\setupPageNumbers
\usepackage{rotating}

\usepackage{lscape}
\usepackage{multirow}
\usepackage{float}
\usepackage{tabularray}
\usepackage{adjustbox}
\usepackage{tabularx,ragged2e,booktabs}
\usepackage{tabulary,booktabs}
\usepackage{tabularray}
\usepackage{academicons}
\usepackage{graphicx} 
\usepackage{lipsum}   

\usepackage{amsmath, amsthm, amssymb, amsfonts}

\usepackage[numbers,sort&compress]{natbib}
\usepackage[utf8]{inputenc}	
\usepackage[T1]{fontenc}	
\usepackage{xcolor}		
\usepackage[colorlinks = true,
            linkcolor = purple,
            urlcolor  = blue,
            citecolor = cyan,
            anchorcolor = black]{hyperref}	
\usepackage{booktabs} 		
\usepackage{nicefrac}		
\usepackage{microtype}		
\usepackage{lineno}		
\usepackage{float}			
\usepackage{lipsum}		

\usepackage{newfloat}
\DeclareFloatingEnvironment[name={Supplementary Figure}]{suppfigure}
\usepackage{sidecap}
\sidecaptionvpos{figure}{c}

\usepackage{titlesec}
\titlespacing\section{0pt}{12pt plus 3pt minus 3pt}{1pt plus 1pt minus 1pt}
\titlespacing\subsection{0pt}{10pt plus 3pt minus 3pt}{1pt plus 1pt minus 1pt}
\titlespacing\subsubsection{0pt}{8pt plus 3pt minus 3pt}{1pt plus 1pt minus 1pt}

\usepackage{tikz,xcolor,hyperref}

\definecolor{lime}{HTML}{A6CE39}
\DeclareRobustCommand{\orcidicon}{
	\begin{tikzpicture}
	\draw[lime, fill=lime] (0,0)
	circle [radius=0.16]
	node[white] {{\fontfamily{qag}\selectfont \tiny ID}};
	\draw[white, fill=white] (-0.0625,0.095)
	circle [radius=0.007];
	\end{tikzpicture}
	\hspace{-2mm}
}
\foreach \x in {A, ..., Z}{\expandafter\xdef\csname orcid\x\endcsname{\noexpand\href{https://orcid.org/\csname orcidauthor\x\endcsname}
			{\noexpand\orcidicon}}
}
\newcommand{\custombio}[3]{
    \begin{minipage}[t]{\columnwidth}
    \vspace{0pt} 
    \parbox[t]{1in}{%
        \vspace{0pt} 
        \includegraphics[width=1in,height=1.25in,clip,keepaspectratio]{#1}%
    }
    \hspace{5mm} 
    \begin{minipage}[t]{0.65\columnwidth}
    \vspace{0pt} 
    {\textsf{\textcolor{blue}{#2}}} 
    {\textsf{\textcolor{black}{#3}}} 
    \end{minipage}
    \end{minipage}
    \vspace{5mm} 
}

\title{BioFusionNet: Deep Learning-Based Survival Risk Stratification in ER+ Breast Cancer Through Multifeature and Multimodal Data Fusion}

\usepackage{authblk}

\author[1]{Raktim Kumar Mondol}
\author[2]{Ewan K.A. Millar}
\author[1]{Arcot Sowmya}
\author[1,*]{Erik Meijering} 
\affil[1]{School of Computer Science and Engineering, University of New South Wales, Sydney, Australia}
\affil[2]{Department of Anatomical Pathology, NSW Health Pathology, St. George Hospital}
\affil[*]{Correspondence: \texttt{erik.meijering@unsw.edu.au}}

\begin{document}

\twocolumn[ 
\begin{@twocolumnfalse} 

\maketitle
\begin{abstract}
Breast cancer is a significant health concern affecting millions of women worldwide. Accurate survival risk stratification plays a crucial role in guiding personalised treatment decisions and improving patient outcomes. Here we present BioFusionNet, a deep learning framework that fuses image-derived features with genetic and clinical data to obtain a holistic profile and achieve survival risk stratification of ER+ breast cancer patients. We employ multiple self-supervised feature extractors (DINO and MoCoV3) pretrained on histopathological patches to capture detailed image features. These features are then fused by a variational autoencoder and fed to a self-attention network generating patient-level features. A co-dual-cross-attention mechanism combines the histopathological features with genetic data, enabling the model to capture the interplay between them. Additionally, clinical data is incorporated using a feed-forward network, further enhancing predictive performance and achieving comprehensive multimodal feature integration. Furthermore, we introduce a weighted Cox loss function, specifically designed to handle imbalanced survival data, which is a common challenge. Our model achieves a mean concordance index of 0.77 and a time-dependent area under the curve of 0.84, outperforming state-of-the-art methods. It predicts risk (high versus low) with prognostic significance for overall survival in univariate analysis (HR=2.99, 95\% CI: 1.88--4.78, p$<$0.005), and maintains independent significance in multivariate analysis incorporating standard clinicopathological variables (HR=2.91, 95\% CI: 1.80--4.68, p$<$0.005).
\end{abstract}
\keywords{Multimodal Fusion\and Breast Cancer\and Whole Slide Images\and Deep Neural Network\and Survival Prediction} 
\vspace{0.35cm}
\end{@twocolumnfalse} 
] 



\section{Introduction}
{B}{reast} cancer poses a significant global health concern, with a high incidence rate and substantial impact on morbidity and mortality~\cite{10.1136/bmjopen-2022-061205, 10.3389/fgene.2021.709027}. The incidence of breast cancer varies across different regions and populations, with higher rates observed in developed countries~\cite{10.1136/bmjopen-2022-061205}. The prevalence of breast cancer in Australia, affecting 1 in 8 women up to the age of 85, is a cause for concern due to its rising incidence rate over the past decade~\cite{10.1111/jan.15094}. This trend highlights the crucial need for accurately predicting survival risks to identify high-risk patients who may benefit from more intensive treatment or monitoring, thereby potentially improving outcomes~\cite{10.3389/fgene.2021.709027}. 

In breast cancer, the estrogen receptor (ER) status plays a critical role in determining treatment strategies and predicting patient prognosis. ER+ breast cancer, which includes Luminal A and Luminal B subtypes, is distinguished by the presence of ERs on cancer cells, making it responsive to hormonal therapies such as tamoxifen~\cite{10.7717/peerj.12202}. 
While Luminal A tumours are usually low-grade with a favorable prognosis (Ki-67$<$14\%), Luminal B tumours are usually higher grade and pose a higher recurrence risk and worse outcome (Ki-67$\geq$14\%)~\cite{10.7717/peerj.12202, 10.1186/s12880-017-0239-z}.  The inherent heterogeneity of breast cancer poses challenges for prediction of prognosis and treatment decisions, particularly in post-menopausal ER+ breast cancer, with previous studies reporting conflicting results on the survival difference between Luminal A and B metastatic breast cancer patients~\cite{10.1186/s12885-022-10098-1, 10.1016/j.breast.2020.09.006}. A common critical clinical dilemma is the selection of those early ER+ breast cancer patients at high risk of recurrence who may benefit from the addition of chemotherapy to endocrine therapy. Therefore, accurate survival risk prediction models specifically tailored for ER+ breast cancer are essential for personalised treatment decisions.

\begin{figure*}[!t]
\centering
\includegraphics[width=\textwidth]{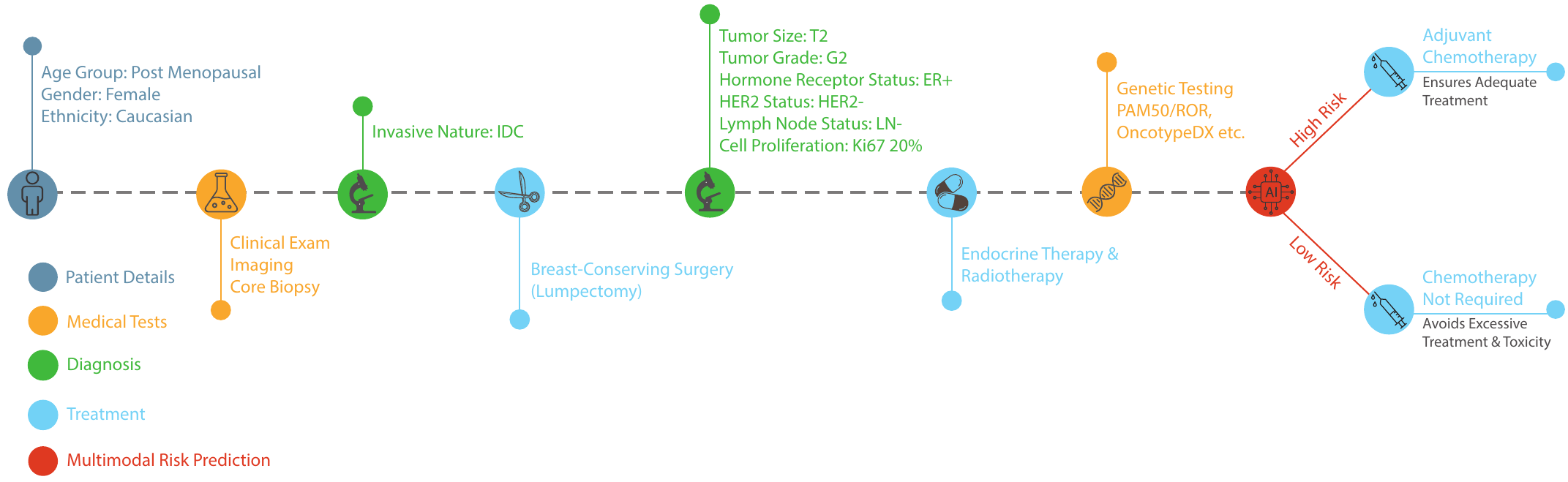}
\smallfigcaption{Illustration of the clinical management pathway in treating breast cancer patients. This example concerns a postmenopausal patient who has been diagnosed with breast cancer, specifically invasive ductal carcinoma (IDC). The process begins with an initial diagnosis through clinical examination, imaging and core biopsy. Following this, surgery is performed to completely excise the tumour, and postoperative tumour histopathological classification is performed to assess key factors including tumour size (e.g. T2), grade (e.g. G2), hormone receptor status (e.g. ER+), HER2 status (e.g. HER2-), lymph node status (e.g. LN-), and proliferation index (e.g. Ki67 20\%). Subsequent treatments may include hormone therapy and radiotherapy. Additional molecular tests, like genetic testing, are utilised to determine specific cancer molecular subtypes and further assess risk of recurrence. The proposed final step in this pathway is the application of our BioFusionNet model. This model combines tumour characteristics, pathology and genetic testing to determine high and low risk patients, thereby guiding personalised treatment decisions and efficiently preventing both under-treatment and over-treatment. For example, low-risk patients might undergo lumpectomy with hormone therapy and radiotherapy, whereas high-risk patients are advised to have chemotherapy in addition to these treatments. Whilst this pathway mirrors current clinical practice, our study streamlines the integration of all available critical data to derive an automated single risk prediction score.}
\label{fig:patientjourney}
\end{figure*}

Traditional methods for survival risk prediction often rely on clinicopathological risk factors (such as age, tumour size, grade, lymph node metastasis and clinical stage),  which may fail to fully capture the complex biology of cancer~\cite{10.7717/peerj.12202, 10.1136/bmjopen-2023-071810, 10.3389/fonc.2021.612450, 10.1016/j.breast.2020.09.006, 10.46754/jssm.2022.12.015}. To address this issue, molecular markers and gene expression profiles have been identified as potential prognostic factors that provide valuable insights into tumour biology and potential therapeutic targets \cite{10.1371/journal.pcbi.1008133, 10.1101/2021.05.26.445720, 10.3390/cancers14081898, 10.1177/15330338221145246, 10.1186/s40164-022-00363-1}. Over the past decade, the integration of genomic testing into treatment decision-making processes has been enhanced by the utilisation of several commercial gene panels, such as Prosigna/PAM50, OncotypeDx and Endopredict among others. In addition, histopathological imaging, which offers in-depth insights into the cellular and tissue characteristics of tumours, plays a vital role in both the diagnosis and prognosis of breast cancer~\cite{10.1007/978-3-030-77211-6_2}.  However, to address the  varied nature of the disease effectively, it is essential to consider all available data modalities.  Therefore, integrating imaging, genetic and clinicopathological information into a single risk prediction model could potentially enhance risk prognostication in the clinic~\cite{10.3389/fonc.2021.612450}. 

In this study, we propose a novel multimodal survival risk prediction model that significantly enhances the prognosis of ER+ breast cancer by integrating histopathology images, genetic profiles, and clinical data. By combining these data modalities, our model captures the complex interplay between cellular, molecular, and clinical factors, thereby improving its predictive accuracy. Moreover, we introduce a weighted Cox loss function specifically designed to handle imbalanced survival data, further enhancing the model's performance. The model is thoroughly evaluated using metrics such as the concordance index (C-index) and time-dependent area under the curve (AUC) score, and its performance is compared to existing methods to establish its efficacy. The model's ability to accurately predict survival risks and categorise patients into distinct risk groups has the potential to inform personalised treatment decisions and improve patient outcomes. Finally, we provide explainable analyses to elucidate the influence of different genes and clinical factors on risk prediction, offering valuable insights into the underlying predictive mechanisms. 
The main contributions of this study are as follows:
\begin{enumerate}
\item A novel multimodal survival risk prediction model integrating histopathology images, genetic profiles, and clinical data for ER+ breast cancer prognosis.
\item A weighted Cox loss function designed to handle imbalanced survival data, improving predictive accuracy.
\item Explainable analyses providing insights into the influence of genes and clinical factors on risk prediction.
\end{enumerate}


\section{Background}
Cancer risk prediction is of paramount importance due to its potential for guiding personalised screening, prevention and treatment strategies, ultimately leading to improved patient outcomes and reduced mortality (Fig.~\ref{fig:patientjourney}). This approach is especially vital in the context of ER+ breast cancer, where accurate risk prediction is essential for identifying individuals at higher recurrence risk. These patients may benefit from more aggressive treatments like chemotherapy  \cite{10.1186/s13058-021-01453-4, 10.1186/s12885-021-08601-1, 10.1200/jco.2005.04.7985, 10.3389/fonc.2021.653243}. On the other hand, risk models are equally critical in recognising lower-risk patients, potentially sparing them from unnecessary treatments and their side effects  \cite{10.1111/cas.14838}. Online algorithms such as Predict\footnote{\url{https://breast.predict.nhs.uk/tool}} and Adjuvant\footnote{\url{https://oncoassist.com/adjuvant-tools/}} are used clinically to estimate the risk of recurrence and the benefit of adding chemotherapy to endocrine therapy, which is a major treatment dilemma.

Developments in deep learning, such as the Cox proportional hazards deep neural network, have  revolutionised cancer research by improving survival data modeling and treatment recommendation systems \cite{10.1186/s12874-018-0482-1}. Multimodal data fusion, which combines information from diverse sources such as imaging, genomics and clinical data, has gained attention in cancer research due to its ability to provide a comprehensive understanding of the disease and improve predictive outcomes \cite{10.3389/fonc.2022.745258, 10.1109/isbi48211.2021.9434033, 10.1038/s43018-022-00416-8}. Cross-attention transformer mechanisms that integrate histopathological images and genomic data capture complementary information from different modalities, leading to improved survival prediction \cite{10.1117/12.2668986}.

In this evolving landscape, models such as MultiDeepCox-SC~\cite{10.1111/cas.15592}, MCAT~\cite{Chen2021}, MultiSurv\cite{multisurv}, HFBSurv~\cite{Li2022}, Pathomic Fusion~\cite{Chen2022}, and TransSurv~\cite{Lv2022} exemplify significant progress in multimodal analysis. These models harness unique strategies to integrate diverse data types for enhanced survival prediction. MultiDeepCox-SC combines histopathological image-derived risk scores with clinical data and gene expression through the Cox proportional hazards (CoxPH) model, providing a detailed risk assessment. MCAT utilises a genomic-guided co-attention layer to map relationships between whole slide images and genomic features, enhancing interpretability in computational pathology. MultiSurv simplifies the integration of multimodal data by merging feature vectors into a unified representation for predicting survival probabilities. In contrast, HFBSurv applies an hierarchical framework with attentional factorised bilinear modules, systematically processing information from lower to higher complexity levels. Pathomic Fusion adopts a gating-based attention mechanism, effectively filtering out noise to focus on salient features across modalities. Finally, TransSurv employs cross-attention transformers to combine histopathological and genomic data, capturing complementary information that significantly improves predictive accuracy.


Despite these advances, the integration of multimodal data, marked by its inherent heterogeneity and dimensional variability, remains a significant challenge, highlighting the need for advanced methodologies to effectively integrate and leverage diverse data sources for robust survival risk assessment \cite{10.1145/3430984.3431030, 10.1145/3394171.3416274, 10.3390/s23031305}. To address this, BioFusionNet introduces a unique strategy by combining self-supervised learning models, specifically DINO and MoCoV3, for feature extraction. This method is further enhanced by a co-dual-cross-attention mechanism for effective multimodal fusion. The co-attention component facilitates synchronised learning from multiple data types by highlighting mutually informative features, whereas the dual cross-attention mechanism provides a deeper interaction layer, allowing for more effective integration of these features. Additionally, BioFusionNet mitigates issues of data imbalance through the implementation of a weighted Cox loss function. This comprehensive approach to multimodal fusion marks a significant advancement in predicting survival risks for ER+ breast cancer, addressing notable gaps in current research.

\begin{figure*}[!t]
\centering
\includegraphics[width=0.9\textwidth]{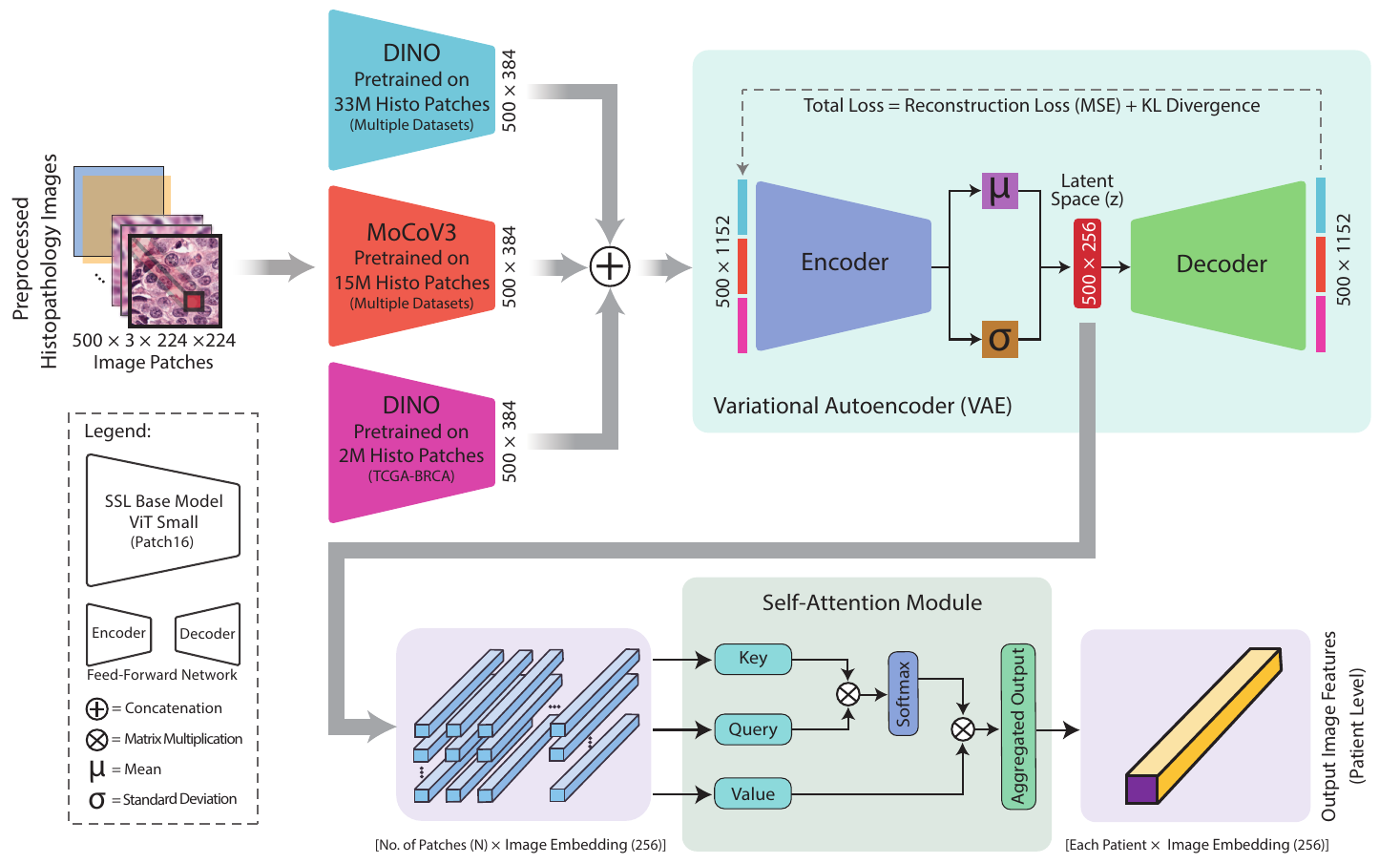}
\smallfigcaption{BioFusionNet Stage 1: The proposed model integrates self-supervised image feature extraction methods, namely DINO and MoCoV3, pretrained on three distinct datasets. Features are concatenated and fed to a Variational AutoEncoder (VAE). Subsequently, the latent space of the VAE is utilised to feed a self-attention network, which aggregates patch-level features into a comprehensive patient-level representation.}
\label{fig:imageprocessing}
\end{figure*}

\subsection{Data Collection}
Our study used hematoxylin-and-eosin-stained (H\&E) formalin-fixed paraffin-embedded (FFPE) digital slides from The Cancer Genome Atlas Breast Invasive Carcinoma (TCGA-BRCA). Whole-slide images (WSIs) from the TCGA-BRCA data collection were downloaded from the \href{https://portal.gdc.cancer.gov/}{GDC Portal} (accessed 25 August 2023). In this work, we chose a subset of 249 cases from the TCGA-BRCA dataset, in order to maintain the proportions of Luminal A and B subtypes as well as the proportion of survival events within each subtype. Among the 249 cases, 83 had survival events, while the remaining cases were censored. The survival events were distributed as follows: 54 events in the Luminal A subtype and 29 events in the Luminal B subtype. For each subtype, we selected cases such that the total was about three times the number of survival events, resulting in 149 Luminal A cases and 100 Luminal B cases. By preserving these proportions, we obtained a subset representative of key characteristics in the full dataset. Additionally, we obtained transcriptome-wide RNA-sequencing data representing mRNA expression levels for a total of 20,438 genes in the reference genome from the TCGA dataset. These data were processed using RNA-sequencing by expectation maximisation (RSEM) and were downloaded from the \href{https://www.cbioportal.org/}{cBioPortal} platform\cite{Li2011}.  This dataset included a range of clinical information for each patient, such as tumour grade, tumour size, lymph node status, age at diagnosis and molecular subtypes. Overall, patients who had WSIs, RNA-sequencing and clinical data available were included in the study. 

\subsection{Data Preparation}

\subsubsection{Slide Annotation}
An expert breast pathologist manually annotated the selected slides using QuPath \cite{Bankhead2017}. The annotation was performed for localisation of the tumour outline, excluding any necrosis but including stroma and tumour infiltrating lymphocytes (TILs). The pathologist was blinded to any molecular or clinical features during annotation.

\subsubsection{Image Data Preparation}
The images were first downsampled to 0.25 µm/pixel, corresponding to approximately 40$\times$ magnification. The annotated tumour regions were processed semi-automatically with QuPath to create 224$\times$224-pixel patches, resulting in approximately 500 nonoverlapping patches per sample. To address staining inconsistencies, vector-based colour normalisation was applied \cite{10.1109/isbi.2009.5193250}.

\subsubsection{RNA-Sequencing Data Preparation}
 From the extensive set of 20,438 genes, we selected genes featured in various commercial assays, namely Oncotype DX, Mammaprint, Prosigna (PAM50), EndoPredict, BCI (Breast Cancer Index), and Mammostrat \cite{10.3390/cancers12051133, 10.3390/curroncol29040213, 10.1007/s10549-020-05688-1, 10.1038/s41416-020-0838-2, 10.1200/go.20.00250, 10.1200/jco.20.00853}, as these are the most relevant genetic markers to our study's objectives. This resulted in a subset of 138 genes. The RNA-sequencing data obtained from the TCGA dataset had already undergone processing using RSEM, and no further normalization was applied to the gene expression values.

\subsubsection{Clinical Data Preparation}
From the clinical data, we selected variables based on their established relevance in breast cancer prognosis and treatment outcomes \cite{10.1002/cncr.20140, 10.1016/s0002-9610(02)00950-9, 10.9734/jammr/2021/v33i2031118, 10.53350/pjmhs2023171656}. Specifically, we included tumour grade (categorised as grade 1\&2 versus grade 3), tumour size ($>$20 mm versus $\leq$20 mm), patient age ($>$55 versus $\leq$55) and lymph node status (positive versus negative). The decision to binarise the clinical data was made to facilitate CoxPH analysis. Binarisation also simplifies both univariate and multivariate hazard analyses and makes it easier to understand how each clinical factor affects the chance of survival.

\begin{figure*}[!t]
\centering
\includegraphics[width=\textwidth]{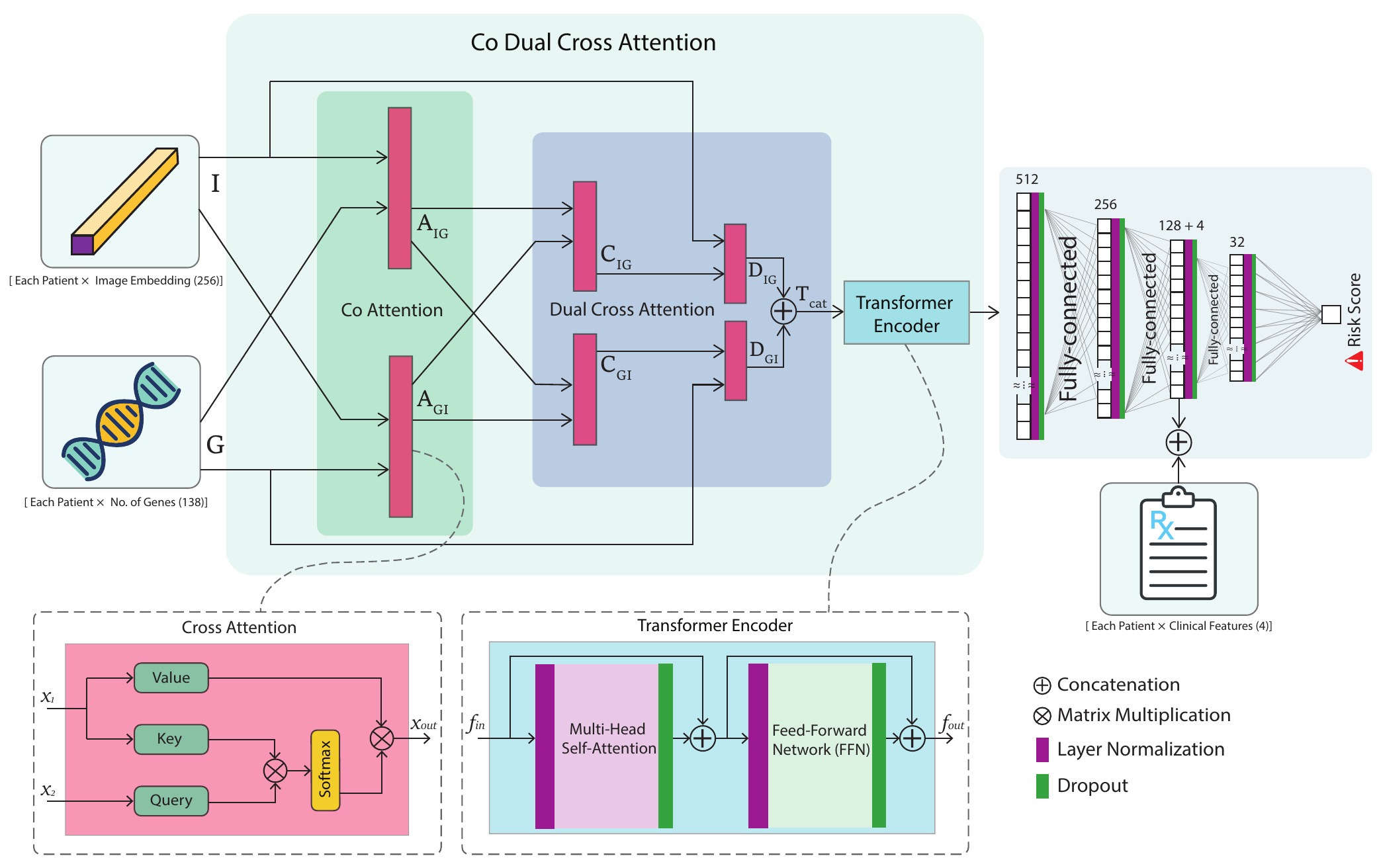}
\smallfigcaption{BioFusionNet Stage 2: The proposed model fuses image embeddings generated from Stage 1 with genetic data through a co-dual-cross-attention mechanism. This fusion is subsequently combined with clinical data using a feed-forward network (FFN), leading to the generation of the final risk score output.}
\label{fig:multimodal}
\end{figure*}
\subsection{Proposed Model}
The proposed deep learning model, which we call BioFusionNet, is an innovative feature extraction and multimodal fusion framework designed to leverage and integrate diverse data types, including histology images, genomic features and clinical data for enhanced cancer outcome prediction (Figs.~\ref{fig:imageprocessing} and \ref{fig:multimodal}). The essence of BioFusionNet lies in its capability to fuse these data modalities into a cohesive tensor representation, effectively capturing both bimodal and trimodal interactions. This approach is aimed at surpassing the performance of traditional unimodal and existing multimodal representations in survival risk prediction.

\subsubsection{Feature Extraction Using DINO and MoCoV3}
Histopathological images, rich in phenotypic information, are pivotal for understanding cancer pathology. BioFusionNet utilises two advanced self-supervised learning models, DINO (self-DIstillation with NO labels) and MoCoV3 (Momentum Contrast version 3), both based on the Vision Transformer (ViT) architecture, to extract morphological features from histology images crucial for identifying cancer-related patterns.

{\bf DINO:} The DINO framework employs a dual-network architecture, consisting of a student and a teacher network, both being ViTs. The student network learns by attempting to replicate the output of the teacher network, which in turn is an exponential moving average of the student's parameters. The core process involves generating multiple augmented views (\(I_1, I_2, \ldots, I_k\)) of a given input image (\(I\)), which are then processed by these networks. The resultant feature vectors from the student (\(F_s\)) and teacher (\(F_t\)) networks are utilised to compute the distillation loss as follows:
\begin{equation}
\mathcal{L}_{\text{D}} = \sum_{i=1}^{k} \text{CrossEntropy}\left(F_s(I_i), \text{Softmax}\left(\frac{F_t(I_i)}{\tau}\right)\right)\!,
\end{equation}
where \( \tau \) represents the temperature scaling parameter. Notably, DINO is pretrained on a broad range of datasets, including BACH, CRC, MHIST, PatchCamelyon and CoNSeP, comprising 33 million patches (DINO33M), and 2 million patches from TCGA-BRCA (DINO2M) \cite{Kang2022, chen2022self}. The output functions for DINO33M and DINO2M, denoted as \( f_{\text{DINO33M}}(x) \) and \( f_{\text{DINO2M}}(x) \) respectively, convert an input image \( x \) of size \( 224 \times 224 \times 3 \) into a \( 1 \times 384 \) feature vector.
We utilised these two models for feature extraction: DINO33M trained on diverse datasets providing a broad perspective and enabling the model to recognise a wide array of general histopathological features, and DINO2M specifically trained on breast cancer data for more specialised and precise feature extraction relevant to breast cancer pathology.

{\bf MoCoV3:} The MoCoV3 framework, which incorporates ViTs, represents a significant advancement in self-supervised learning through its adoption of the momentum contrast (MoCo) approach~\cite{mocoV3}. At the heart of this framework lies the contrastive learning mechanism, designed to differentiate between positive and negative pairs, thereby enhancing the model's feature learning capabilities. MoCoV3's architecture is defined by two main components: a query encoder that processes the current batch of images, and a key encoder updated via a momentum-based moving average of the query encoder's parameters:
\begin{equation}
\theta_k \leftarrow m \theta_k + (1 - m) \theta_q,
\end{equation}
where \( \theta_k \) and \( \theta_q \) are the parameters of the key and query encoders respectively and \( m \) is the momentum coefficient. This enables the key encoder to maintain a queue of encoded keys representing previously seen images, thus enhancing the model's ability to maximise agreement between differently augmented views of the same image (positive pairs) and minimise similarity with other images (negative pairs). The framework uses the InfoNCE loss~\cite{Oord2018}:
\begin{equation}
\mathcal{L}_{\text{InfoNCE}} = -\log \frac{\exp(q \cdot k^+ / \tau)}{\sum_{i=0}^{K} \exp(q \cdot k_i / \tau)},
\end{equation}
where \(q\) and \(k^+\) are the query and positive key feature vectors, \(k_i\) are the negative key vectors, \(K\) is the number of negative keys and \(\tau\) is the temperature parameter. MoCoV3 has been pretrained on an extensive collection of 15 million histology patches from over 30 thousand WSIs derived from the TCGA and Pathology AI Platform (PAIP) datasets, encompassing a wide variety of cancer types and histological features. This endows the model with a robust and versatile capability to extract meaningful features from histopathological data \cite{Wang2022}. Similar to DINO, the output function of MoCoV3, \( f_{\text{MoCoV3}}(x) \), transforms an input image \( x \) of dimensions \( 224 \times 224 \times 3 \) into a \( 1 \times 384 \) feature vector.

\subsubsection{Unimodal Feature Integration}
Extracted features from DINO33M, DINO2M and MoCoV3 are concatenated to form a comprehensive \( 1 \times 1152 \) feature vector $f_{\text{cat}}(x) = f_{\text{DINO33M}}(x) \oplus f_{\text{DINO2m}}(x) \oplus f_{\text{MoCoV3}}(x)$ (Fig.~\ref{fig:imageprocessing}).
Following this, we employ a VAE to encode the integrated image features into a 256-dimensional feature in latent space. The latent feature vector is then passed through a self-attention model and aggregated using sum pooling to generate patient-level features.

\subsubsection{Feature Fusion Using Variational Autoencoding}
Our VAE consists of an encoder and a decoder. The encoder function \( f_{\text{enc}} \) maps the concatenated feature vector \( f_{\text{cat}}(x) \) to the latent space by generating the mean \( \mu \) and standard deviation \( \sigma \) of the latent representation:
\begin{equation}
(\mu, \sigma) = f_{\text{enc}}(f_{\text{cat}}(x)).
\end{equation}
With 500 patches per patient, the latent space is structured as a matrix of size \( 500 \times 256 \). This is achieved by sampling \( z \) using the reparameterization trick:
\begin{equation}
z = \mu + \sigma \cdot \epsilon, \quad \epsilon \sim \mathcal{N}(0, I), \quad z \in \mathbb{R}^{500 \times 256}.
\end{equation}
The decoder \( f_{\text{dec}} \) then attempts to reconstruct the input from the latent variable \( z \):
\begin{equation}
\hat{x} = f_{\text{dec}}(z).
\end{equation}
The VAE is optimised using a loss function that combines mean squared error (MSE) for reconstruction accuracy and Kullback-Leibler (KL) divergence for distribution regularisation:
\begin{equation}
\mathcal{L}_{\text{VAE}} = \text{MSE}(\hat{x}, x) + \beta \cdot \text{KL}(\mathcal{N}(\mu, \sigma^2) \| \mathcal{N}(0, I)),
\end{equation}
where \(\beta\) balances the reconstruction and regularisation terms. The MSE term ensures that the reconstructed image closely resembles the original input, while the KL divergence term encourages the latent distribution to approximate a standard normal distribution. This enables BioFusionNet to effectively blend the features from the different self-supervised models, enhancing the overall feature representation.

\subsubsection{Patch-to-Patient Aggregation}
To aggregate the patch-level features from the latent space of the VAE into a comprehensive patient-level representation capturing the interdependencies among image patches, our model uses self-attention (Fig.~\ref{fig:imageprocessing}). The self-attention module computes a weighted sum of the key (\(K\)), query (\(Q\)), and value (\(V\)) vectors:
\begin{equation}
y_i = \sum_{j=1}^{500} \text{Softmax}\left(s(Q_i,K_j)\right)V_j,
\end{equation}
where \( s(Q_i,K_j) \) is the attention function that determines the relevance between each query and key pair. This offers two significant benefits. First, by leveraging the VAE's latent vectors, the model focusses on the most pertinent image features, resulting in a more precise feature representation. Second, by considering all latent representations, the self-attention mechanism contextualises each patch within the broader histopathology of the patient. This aggregation process effectively combines the patch-level features of all 500 patches, taking into account their relevance and interdependencies, to generate a comprehensive patient-level representation.

\subsubsection{Multimodal Fusion Using Co Dual Cross Attention}
To integrate patient-level image embeddings with genetic feature data, our model uses a co-dual-cross-attention mechanism (Fig.~\ref{fig:multimodal}). This is achieved using a complex architecture comprising co-attention and dual-cross-attention modules.

The co-attention module applies linear transformations to the image embeddings \(I\) and genetic features \(G\), yielding their respective query (\(Q\)), key (\(K\)), and value (\(V\)) vectors, and computes co-attention scores scaled by \(\sqrt{d_k}\), where \(d_k\) is the dimensionality of the keys:
\begin{align}
A_{IG} &= \text{Softmax}\left(Q_I K_G^T / \sqrt{d_k}\right) V_G,\\
A_{GI} &= \text{Softmax}\left(Q_G K_I^T / \sqrt{d_k}\right) V_I,
\end{align}
where \(A_{IG}\) represents the attention from images to genetic features, and \(A_{GI}\) the attention from genetic features to images. The Softmax function normalises these scores, facilitating an effective weighting of feature importance in the fusion process. This bidirectional attention prepares the ground for more complex interactions in the subsequent stage of the co-dual-cross-attention mechanism.
\begin{table*}[!t]
\centering
\caption{Performance comparison of multimodal and unimodal models for cancer risk prediction using C-index.}
\label{tab:result1}
\resizebox{\textwidth}{!}{%
\begin{tabular}{cccccccc}
\toprule
\multirow{2.5}{*}{Fold} & Imaging+Genetic+Clinical & Imaging+Genetic & Imaging & \multicolumn{2}{c}{Clinical} & \multicolumn{2}{c}{{Genetic}} \\
\cmidrule(r){2-2} \cmidrule(r){3-3} \cmidrule(r){4-4} \cmidrule(r){5-6} \cmidrule(r){7-8}
& BioFusionNet & BioFusionNet & BioFusionNet & CoxPH & MLP & CoxPH & MLP \\
\midrule
1 & 0.78 & 0.73 & 0.58 & 0.52 & 0.66 & 0.62 & 0.66 \\
2 & 0.71 & 0.72 & 0.69 & 0.42 & 0.64 & 0.55 & 0.73 \\
3 & 0.72 & 0.69 & 0.61 & 0.59 & 0.69 & 0.51 & 0.66 \\
4 & 0.81 & 0.75 & 0.70 & 0.64 & 0.70 & 0.62 & 0.75 \\
5 & 0.82 & 0.65 & 0.69 & 0.72 & 0.66 & 0.63 & 0.67 \\
\midrule
Mean $\pm$ Std & $0.77\pm 0.05$ & $0.71\pm 0.04$ & $0.65\pm 0.05$ & $0.58\pm0.11$ & $0.67\pm0.02$ & $0.59\pm0.05$ & $0.69\pm0.04$ \\
\bottomrule
\end{tabular}%
}
\end{table*}
Subsequently, the dual-cross-attention module further refines the integration of the image and genetic features in two distinct stages. The first stage concerns the interaction between the co-attended image features (\(A_{IG}\)) and co-attended genetic features (\(A_{GI}\)). The cross attention is computed as:
\begin{align}
C_{IG} &= \text{Softmax}\left(A_{IG} (A_{GI})^T / \sqrt{d_k}\right) A_{IG},\\
C_{GI} &= \text{Softmax}\left(A_{GI} (A_{IG})^T / \sqrt{d_k}\right) A_{GI},
\end{align}
where \(C_{IG}\) represents the cross-attention output when image features attend to genetic features, and \(C_{GI}\) the reverse. This stage is crucial for enhancing each modality by integrating contextually relevant information from the other. In the second stage, the outputs from the first stage are further refined by reapplying them to their respective original features. This enhances the depth of the multimodal integration:
\begin{align}
D_{IG} &= \text{Softmax}\left(C_{IG} I^T / \sqrt{d_k}\right) I,\\
D_{GI} &= \text{Softmax}\left(C_{GI} G^T / \sqrt{d_k}\right) G,
\end{align}
where \(D_{IG}\) and \(D_{GI}\) denote the refined cross-attention outputs, further enhancing the original image and genetic data with additional contextual insights.
\begin{algorithm}[!t]
\caption{Weighted Cox Loss}
\label{alg:weighted_cox_loss} 
\begin{algorithmic}[1]
\Require
$r$: risks (log hazard ratios), $e$: events, $w$: weights
\Ensure
$\mathcal{L}_\text{WCox}$: negative log likelihood loss
\State Compute $E_w = \sum_{i=1}^{N} w_i e_i$ as total weighted events
\State Sort samples by descending $r$ and align $e$, $w$
\For{$i \in {1, \dotsc, N}$}  \Comment{N is the number of samples}
    \State Compute hazard ratio $h_i = \exp(r_i)$ 
\EndFor
\State Init weighted cumulative hazard $H_{w_0} = 0$
\For{$i \in {1, \dotsc, N}$}
    \State Update cumulative sum $H_{w_i} = H_{w_{i-1}} + w_i h_i$
\EndFor
\For{$i \in {1, \dotsc, N}$}
\State Compute uncensored log likelihood \par $u_i = w_i (r_i - \log(H_{w_i}))$
\EndFor
\State Compute $c = u \odot e$ \Comment{$\odot$ is the element-wise product}
\State Compute loss $\mathcal{L}_\text{WCox} = - \frac{1}{E_w} \sum_{i=1}^{N} c_i$\\
\Return $\mathcal{L}_\text{WCox}$
\end{algorithmic}
\end{algorithm}
By sequentially processing the attended features, the model achieves a richer and more contextually informed representation of the fused image and genetic data, suited for complex tasks like risk prediction. The concatenated output $T_{\text{cat}} = D_{IG} \oplus D_{GI}$ is fed to a Transformer Encoder, which employs multiple layers of self-attention and a feed-forward network (FFN) to achieve a deeper assimilation and transformation of the fused features. The Transformer Encoder consists of $N$ identical layers, each containing two sublayers: a multihead self-attention mechanism and a position-wise fully connected FFN. In the multihead self-attention sublayer, the input is first linearly projected into a number of different subspaces equal to the number of attention heads. The self-attention mechanism is then applied to each subspace independently, allowing the model to jointly attend to information from different representation subspaces at different positions. The outputs from all heads are then concatenated and linearly projected back to the original dimension. The second sublayer is a position-wise FFN, which consists of two linear transformations with a ReLU activation in between:
\begin{equation}
\text{FFN}(x) = \max(0, xW_1 + b_1)W_2 + b_2
\end{equation} 
where $W_1$, $W_2$, $b_1$, and $b_2$ are learnable parameters. A residual connection is employed around each of the two sublayers, followed by layer normalisation. The output of the Transformer Encoder represents the deeply integrated features from both the imaging and genetic modalities and is further processed by four fully-connected layers, the third of which also integrates the clinical information (Fig.~\ref{fig:multimodal}). The integration of clinical variables at this stage is crucial due to their dimensionality and characteristics. Clinical data, comprising only four features, may be overshadowed by the higher-dimensional features from histopathological images and genetic profiles if introduced earlier in the model. By integrating these clinical variables in a later layer of the network, closer to the output, their impact is more effectively mapped onto the model's survival risk prediction. This approach, referred to as `late fusion', allows the clinical variables to have a more significant influence on the final output compared to `early fusion', where clinical variables are integrated in earlier layers of the network.

This multimodal integration results in a holistic representation of both the phenotypic and genotypic information of ER+ breast cancer. Finally, the network employs a linear output layer that predicts the survival risk score.

\subsection{Proposed Loss Function}
For training BioFusionNet, we propose a novel loss function termed the weighted Cox loss (computed by Algorithm~\ref{alg:weighted_cox_loss}), which is tailored to address the challenges of imbalanced survival data (a common issue in survival analysis):
\begin{equation}
\label{wloss}
\mathcal{L}_{\text{WCox}} = - \frac{1}{\sum_{i=1}^{N} w_i e_i} \sum_{i=1}^{N} w_i e_i (r_i - \log(H_{w_i})),
\end{equation}
where \(e_i\) denotes the event occurrence, \(w_i\) the assigned weight, \(r_i\) the log hazard ratio, \(H_{w_i}\) the weighted cumulative hazard and $N$ the number of samples. Unlike the traditional Cox proportional hazards loss ($\mathcal{L}_{\text{Cox}}$) \cite{10.1186/s12874-018-0482-1}, the proposed loss uses weighting to mitigate the effects of uneven distribution of events within the dataset. In this work we used \( w_i = 3 \), considering that censored data (denoted as `0') is almost three times as prevalent as event data (denoted as `1'), and thus the sensitivity of the  loss function to the latter should be enhanced accordingly, mitigating the bias towards censored data.

\subsection{Model Training}
The training of BioFusionNet is divided into two distinct stages as follows (Figs.~\ref{fig:imageprocessing} and \ref{fig:multimodal}):

\textbf{Stage 1: Feature Extraction:} The self-supervised pretrained models DINO33M, MoCoV3 and DINO2M extract features from histopathology image patches, which are concatenated and then fed into a VAE to produce embeddings, which in turn are processed by a self-attention module to produce a patient-level feature vector (Fig.~\ref{fig:imageprocessing}). The VAE was optimised using AdamW with a learning rate of 0.0001 and a batch size of 12. The employed loss function is a combination of MSE and KL divergence, targetting the construction of an advanced latent space to generate detailed patient-level features.

\textbf{Stage 2: Risk Prediction:} The proposed co-dual-cross-attention mechanism, followed by multiple FFNs and a final output node that uses a linear activation function, predicts the patient-level risk from the image-based, genetic and clinical information (Fig.~\ref{fig:multimodal}). Here, training was performed using the proposed weighted Cox loss $(\mathcal{L}_\text{WCox})$, which was optimised using the Adam algorithm with a learning rate of 0.001 and a batch size of 12. To mitigate overfitting, an early stopping mechanism based on the validation loss was implemented. This involved halting the training process after a patience period of 10 epochs if no improvement was observed.

This two-stage training approach for BioFusionNet was motivated by two key factors. First, the computational complexity of training sophisticated models, especially those incorporating elements like multiple feature extractors, VAE, and various attention blocks, can be significantly high. An end-to-end training process often requires increased memory usage and even more data, which may not be feasible always. By adopting a two-stage training process, we effectively manage this complexity, breaking down the training into more manageable parts. Second, a modular design facilitates conducting ablation experiments. By training in stages, we can isolate the effects of specific modules or features, systematically analysing their impact and optimising their configuration. This modular approach enhances our ability to refine and improve the learning architecture.

Both training stages used a dataset comprising 199 training samples and 50 validation samples within a five-fold cross-validation framework. The trained model predicts a continuous risk score for every patient within each validation fold. For survival analysis, we employed the median risk score $\theta_{\text{opt}}$, derived from each training set, as a threshold to classify patients in the validation set into two categories: high risk (risk score $>$ $\theta_{\text{opt}}$) and low risk (risk score $<$ $\theta_{\text{opt}}$). We note that $\theta_\text{opt}$ is not a user-defined parameter, but its value depends on and is automatically calculated from the specific training set used in each fold, and thus varies depending on the dataset. To achieve optimal performance, we used Optuna\footnote{\url{https://optuna.org/}} for hyperparameter optimisation, tuning key parameters including learning rate, weight decay, number of neurons, number of layers, and dropout for each model.

\subsection{Evaluation Metrics}
To quantitatively evaluate survival risk score prediction, we employed the concordance index (C-index) and the area under the curve (AUC) as our primary metrics. The C-index assesses the concordance between predicted survival times and observed outcomes, especially in the presence of censored data:
\begin{equation}
\label{tab:cindex}
\text{C-index} = \frac{\sum\limits_{i=1}^{n} \sum\limits_{j=1}^{n} \mathcal{I}(y_i < y_j, \delta_i = 1) \mathcal{I}(\hat{f}(x_i) < \hat{f}(x_j))}{\sum\limits_{i=1}^{n} \sum\limits_{j=1}^{n} \mathcal{I}(y_i < y_j, \delta_i = 1)}.
\end{equation}
where $n$ is the number of patients, $y_i$ and $y_j$ denote the observed survival times, $\delta_i$ indicates whether the event was observed (not censored), $\hat{f}(x_i)$ represents the predicted risk for the $i$th patient, and $\mathcal{I}$ is the indicator function that returns 1 when its condition is true and 0 otherwise. The time-dependent AUC offers a dynamic view of the model accuracy over time $t$ and incorporates weights $\omega_i$, and is calculated using the following formula:
\begin{equation}
\label{tab:auc}
\text{AUC}(t) = 
\frac{\sum\limits_{i=1}^{n}\sum\limits_{j=1}^{n} \omega_i \mathcal{I}(y_i \leq t) \mathcal{I}(y_j > t) \mathcal{I}(\hat{f}({x}_j) \leq \hat{f}({x}_i))}{\sum\limits_{i=1}^{n} \omega_i \mathcal{I}(y_i \leq t) \sum\limits_{i=1}^{n} \mathcal{I}(y_i > t)}.
\end{equation} 
Both the C-index and AUC values range from 0 to 1, with higher values indicating better performance.
In addition to these metrics, we also compared the computational properties of the models in terms of the number of trainable parameters, memory usage, and floating-point operations (FLOPS).
\begin{table}[!t]
\caption{Performance comparison of multimodal fusion methods for cancer risk prediction.}
\label{tab:fusionmethodresult}
\resizebox{1\columnwidth}{!}{%
\begin{tabular}{lcccccc}
\toprule
\multirow{2}{*}{Method} & \multirow{2}{*}{Fold} & \multicolumn{2}{c}{C-index} & \multicolumn{2}{c}{AUC} \\
\cmidrule(lr){3-4} \cmidrule(lr){5-6}
& & Value & Mean $\pm$ Std & Value & Mean $\pm$ Std \\
\midrule
\multirow{5}{*}{MultiSurv \cite{multisurv, Venugopalan2021, Steyaert2023}} & 1 & 0.71 & \multirow{5}{*}{$0.63\pm0.07$} & 0.74 & \multirow{5}{*}{$0.63\pm0.09$} \\
& 2 & 0.59 & & 0.52 & \\
& 3 & 0.60 & & 0.61 & \\
& 4 & 0.69 & & 0.70 & \\
& 5 & 0.54 & & 0.57 & \\
\midrule
{\multirow{5}{*}{MultiDeepCox-SC \cite{10.1111/cas.15592}}} & 1 & 0.71 & { \multirow{5}{*}{$0.60\pm0.08$}} & 0.68 & {\multirow{5}{*}{$0.58\pm0.08$}} \\
{} & 2 & 0.68 & {} & 0.66 & {} \\
{} & 3 & 0.55 & {} & 0.49 & {} \\
{} & 4 & 0.58 & {} & 0.57 & {} \\
{} & 5 & 0.50 & {} & 0.50 & {} \\
\midrule
\multirow{5}{*}{HFBSurv \cite{Li2022}} & 1 & 0.58 & \multirow{5}{*}{$0.54\pm0.07$} & 0.51 & \multirow{5}{*}{$0.49\pm0.04$} \\
& 2 & 0.47 & & 0.51 & \\
& 3 & 0.56 & & 0.45 & \\
& 4 & 0.45 & & 0.44 & \\
& 5 & 0.62 & & 0.53 & \\
\midrule
\multirow{5}{*}{PathomicFusion \cite{Chen2022}} & 1 & 0.63 & \multirow{5}{*}{$0.52\pm0.08$} & 0.68 & \multirow{5}{*}{$0.47\pm0.23$} \\
& 2 & 0.43 & & 0.10 & \\
& 3 & 0.56 & & 0.54 & \\
& 4 & 0.50 & & 0.63 & \\
& 5 & 0.46 & & 0.38 & \\
\midrule
\multirow{5}{*}{MCAT \cite{Chen2021}} & 1 & 0.71 & \multirow{5}{*}{$0.70\pm0.04$} & 0.70 & \multirow{5}{*}{$0.71\pm0.04$} \\
& 2 & 0.69 & & 0.67 & \\
& 3 & 0.64 & & 0.65 & \\
& 4 & 0.70 & & 0.69 & \\
& 5 & 0.76 & & 0.72 & \\
\midrule
\multirow{5}{*}{TransSurv \cite{Lv2022}\cite{10.1117/12.2668986}} & 1 & 0.70 & \multirow{5}{*}{$0.69\pm0.04$} & 0.68 & \multirow{5}{*}{$0.66\pm0.04$} \\
& 2 & 0.61 & & 0.60 & \\
& 3 & 0.69 & & 0.65 & \\
& 4 & 0.69 & & 0.67 & \\
& 5 & 0.74 & & 0.72 & \\
\midrule
\multirow{5}{*}{BioFusionNet (Proposed)} & 1 & 0.78 & \multirow{5}{*}{$0.77\pm0.05$} & 0.82 & \multirow{5}{*}{$0.84\pm0.05$} \\
& 2 & 0.71 & & 0.93 & \\
& 3 & 0.72 & & 0.79 & \\
& 4 & 0.81 & & 0.81 & \\
& 5 & 0.82 & & 0.83 & \\
\bottomrule
\end{tabular}%
}
\end{table}

\begin{figure*}[t]
\centering
\begin{tabular}{@{}ccc@{}}
    \includegraphics[width=0.32\textwidth]{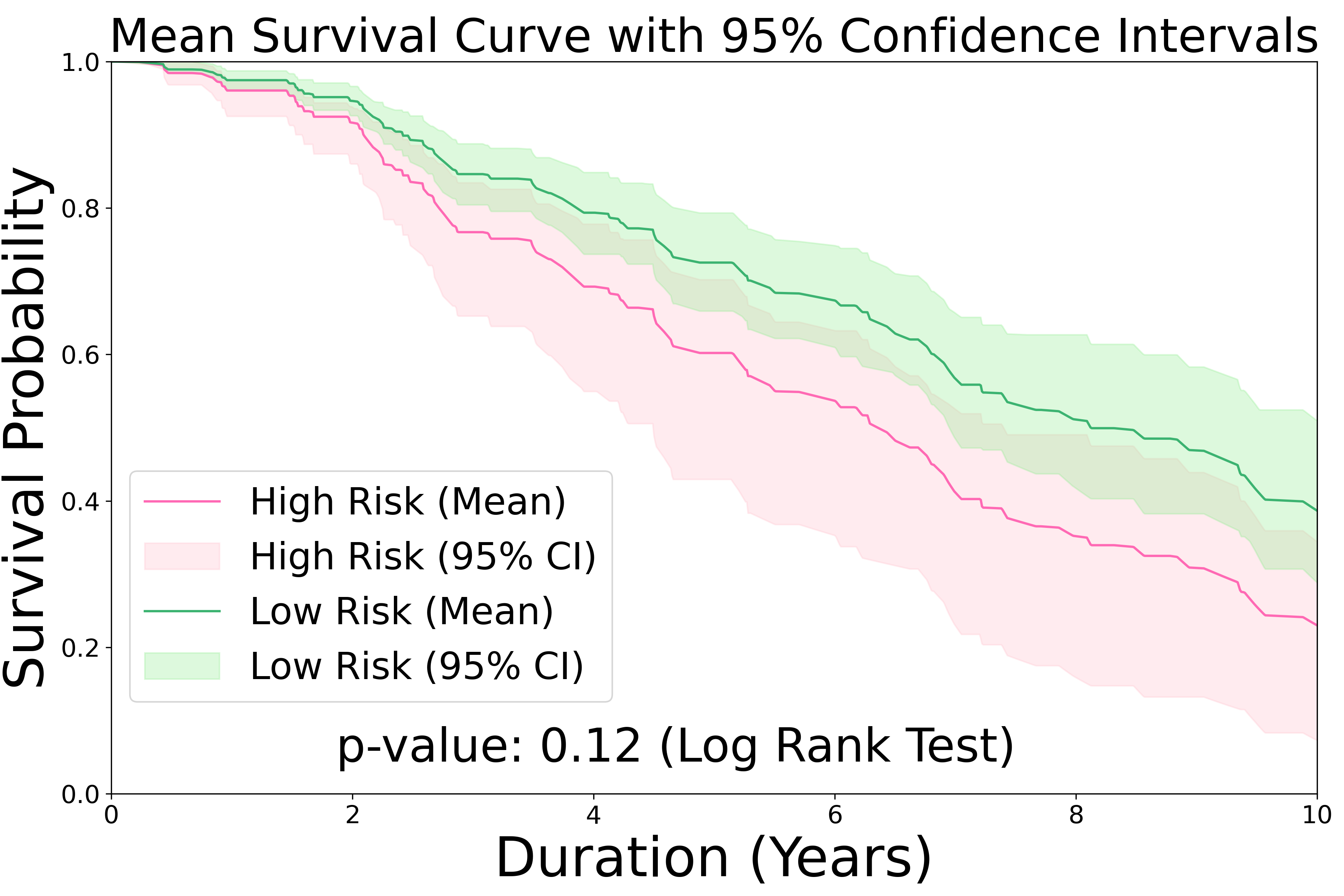} &
    \includegraphics[width=0.32\textwidth]{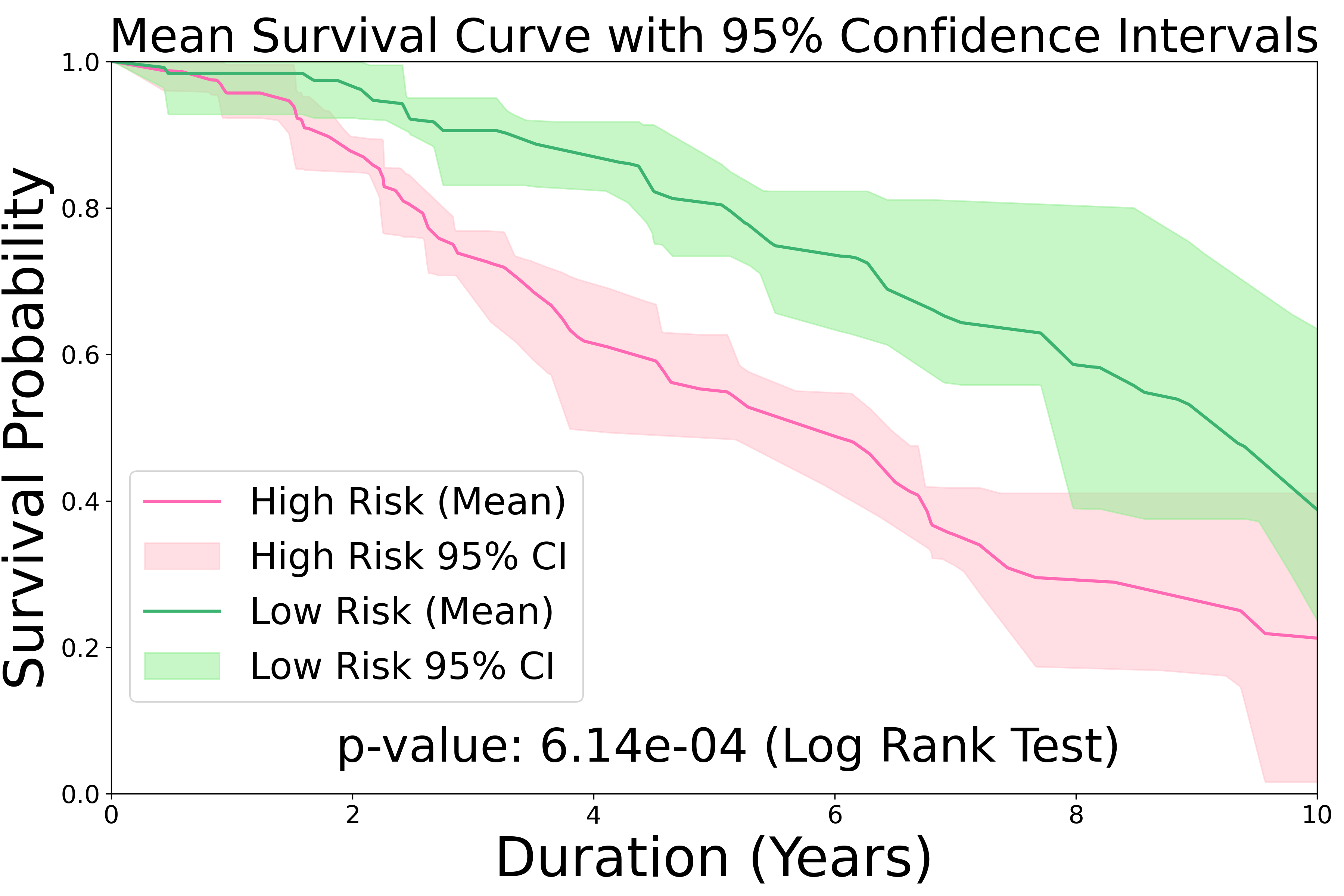} &
    \includegraphics[width=0.32\textwidth]{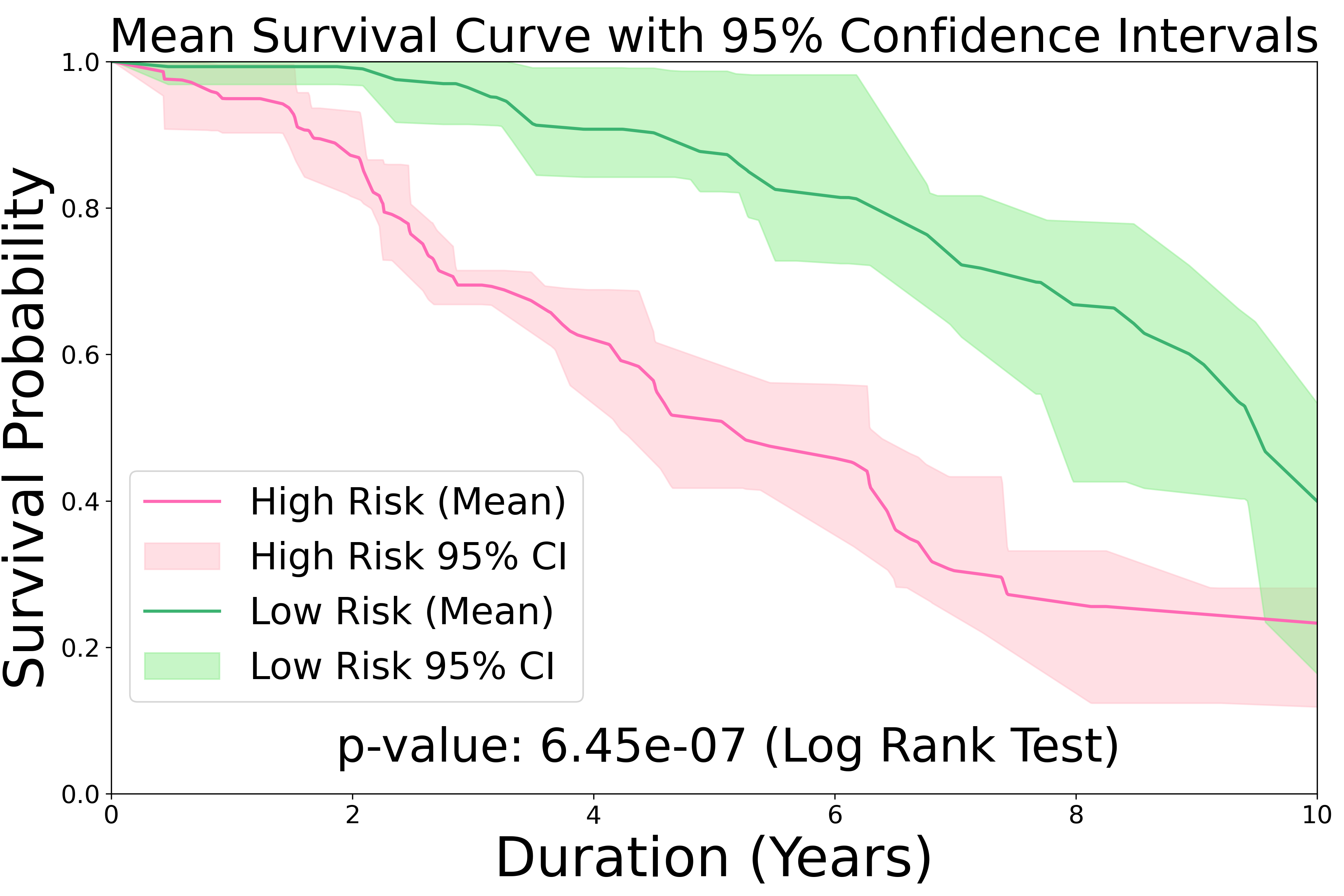} \\
    (a) CoxPH (Clinical) &
    (b) MCAT (Multimodal) &
    (c) BioFusionNet (Multimodal) \\[1em]
    \includegraphics[width=0.32\textwidth]{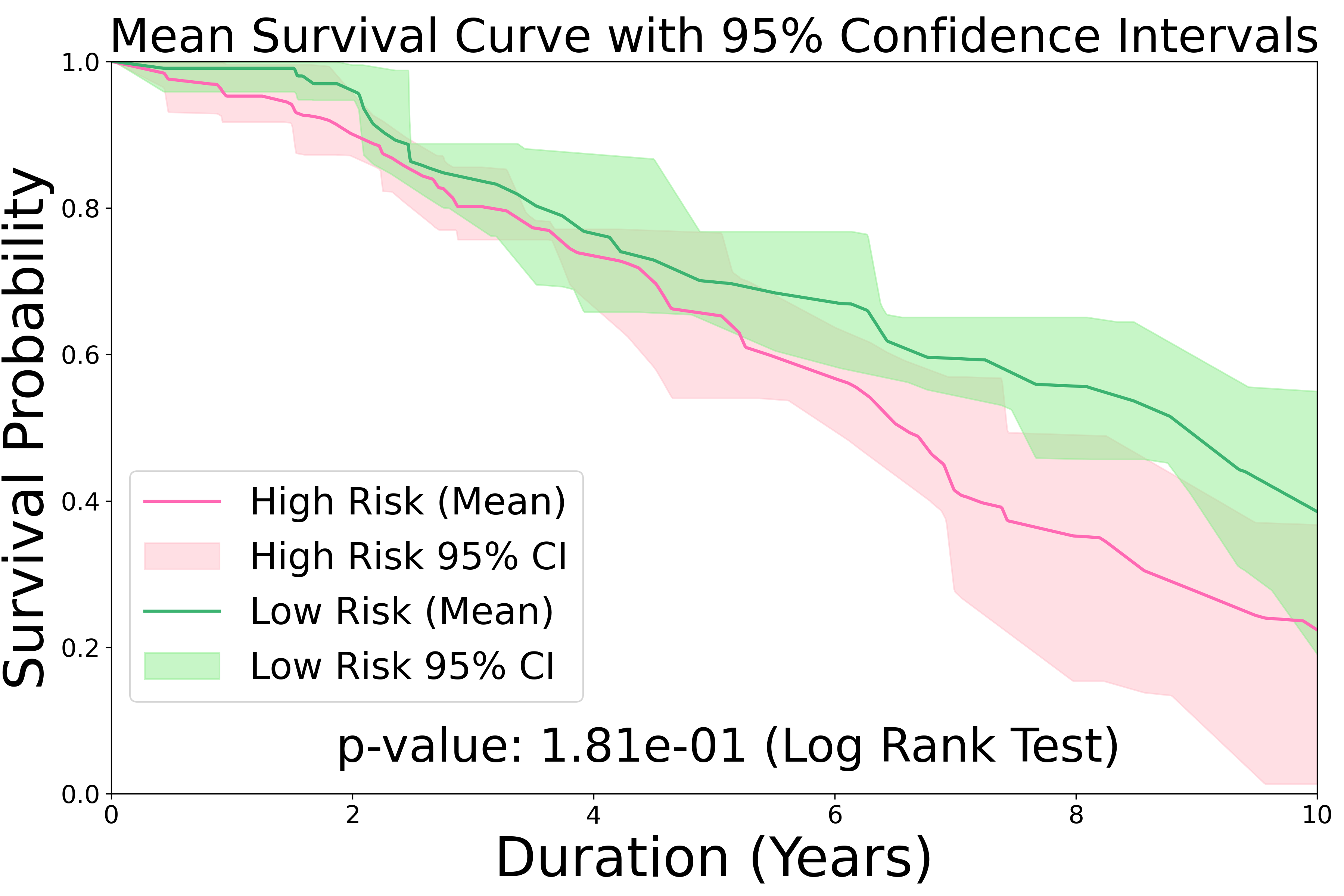} &
    \includegraphics[width=0.32\textwidth]{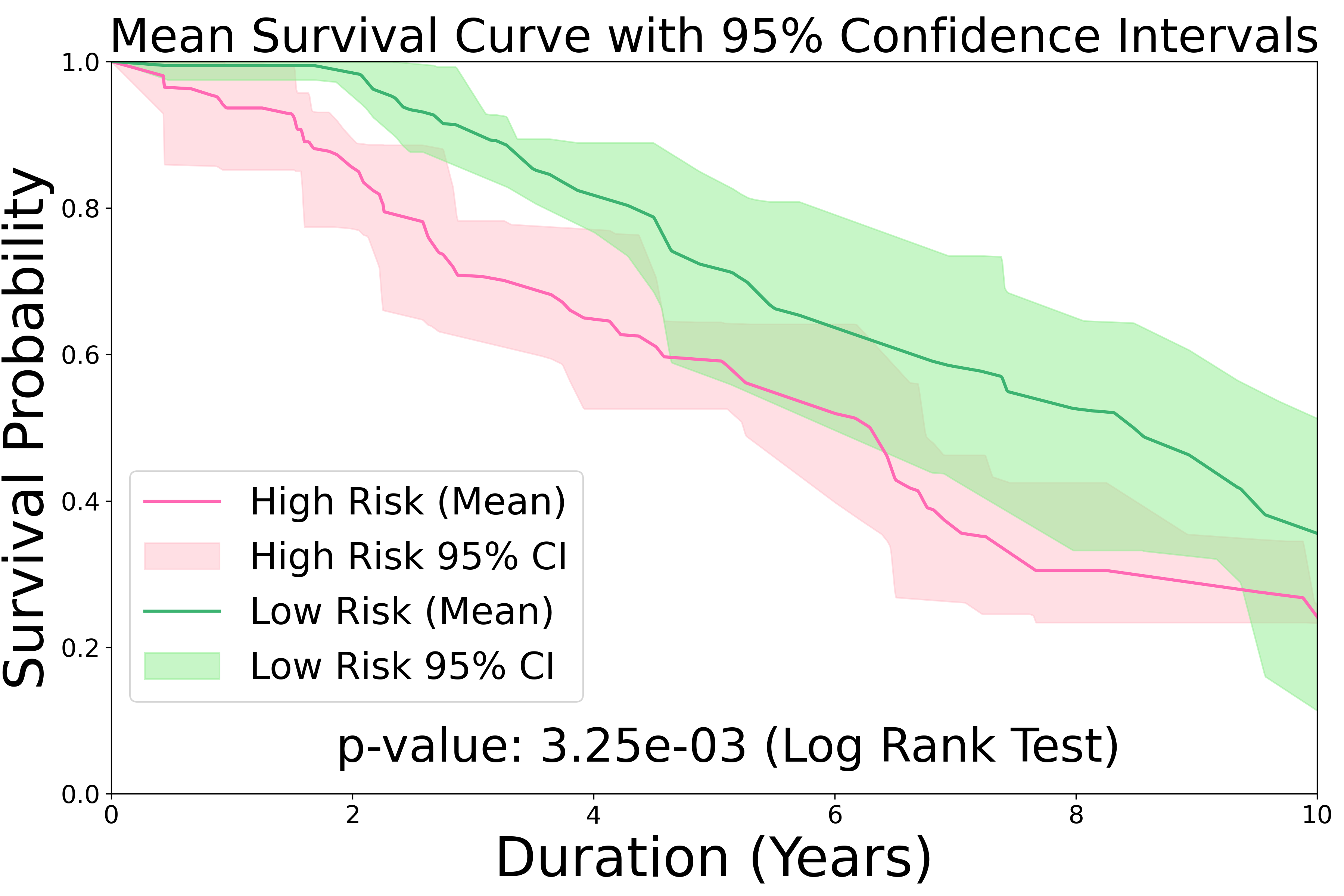} &
    \includegraphics[width=0.32\textwidth]{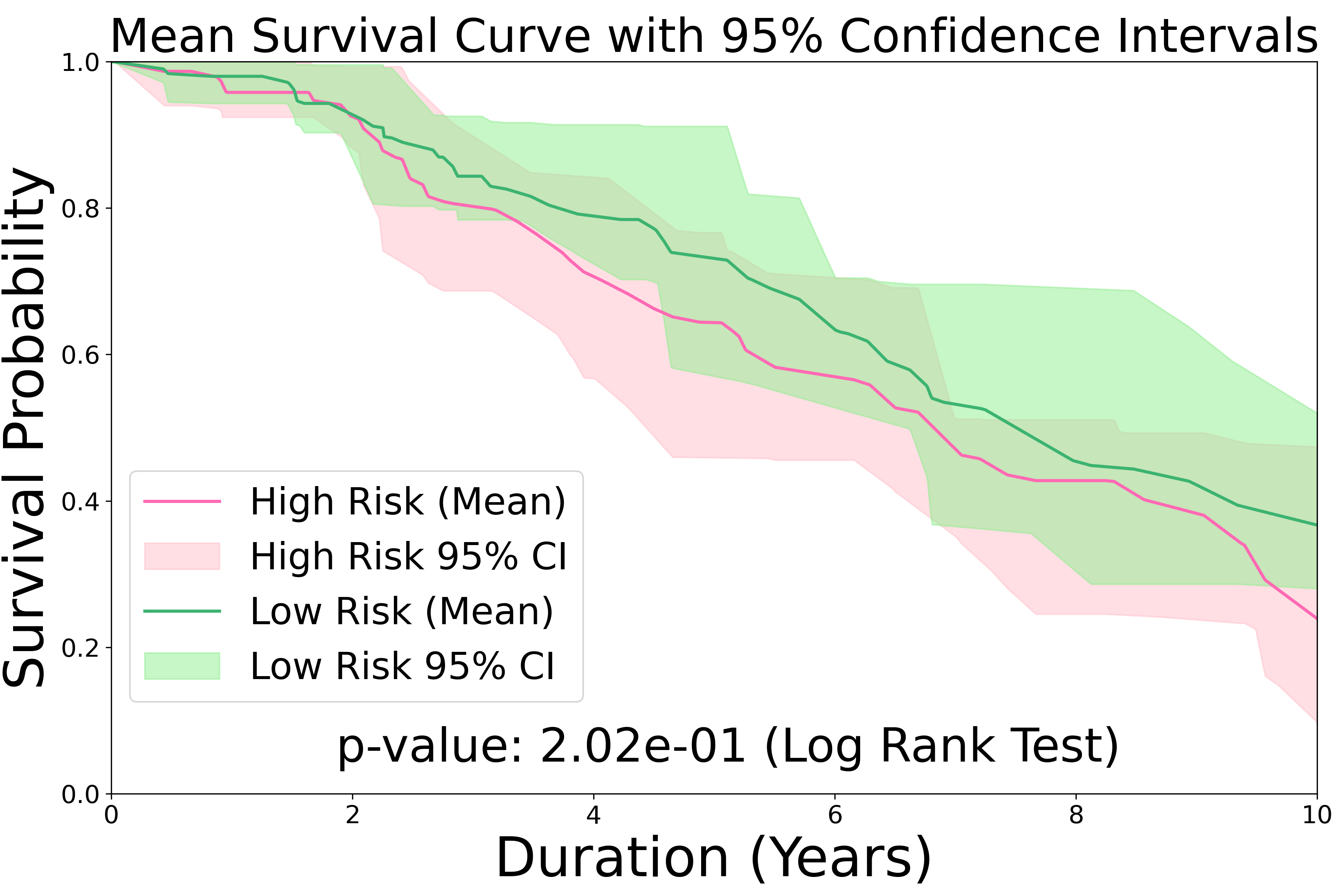} \\
    (d) HFBSurv (Multimodal) &
    (e) MultiSurv (Multimodal) &
    (f) PathomicFusion (Multimodal) \\[0.5em]
\end{tabular}
\smallfigcaption{Performance comparison of BioFusionNet and other methods using Kaplan-Meier survival curves.}
\label{fig:kmplot}
\end{figure*}

\section{Experimental Results}

\subsection{Comparison Across Modalities}
The effectiveness of BioFusionNet in cancer risk prediction was evaluated by comparing its C-index performance across different modality configurations. The results (Table \ref{tab:result1}) show that the mean performance in the cross-validation experiments consistently increased from using only imaging data, to using imaging and genetic data, to combining imaging, genetic and clinical data. Two traditional methods, specifically CoxPH and the MultiLayer Perceptron (MLP), were also evaluated and showed no consistent advantage when using only clinical or only genetic data, and their performance was inferior to that of BioFusionNet. This suggests that BioFusionNet's architecture and training process allow it to effectively extract and integrate complementary information from the three different modalities, leading to improved predictive performance, although it remains elusive exactly how the information in each modality complements the information in the other.

\subsection{Comparison With State-of-the-Art Fusion Methods}
The performance of BioFusionNet was compared with several state-of-the-art multimodal fusion methods for cancer risk prediction in terms of both C-index and AUC (Table \ref{tab:fusionmethodresult}). For this experiment we included methods using concatenation (MultiSurv), Cox proportional hazards with image-derived risk scores (MultiDeepCox-SC), hierarchical attention (HFBSurv), gating attention (PathomicFusion), co-attention (MCAT) and cross attention (TransSurv). To ensure fair comparison, we used the same hyperparameter optimisation framework (Optuna) for all models. The AUC was calculated using average values over 5-year and 10-year periods. The proposed model consistently outperformed all these previous methods, showing substantial improvements in both metrics. However, BioFusionNet is computationally demanding, requiring 12.11 G (Giga) FLOPS, the highest among the compared models, although it is more memory efficient than PathomicFusion (Table~\ref{model_complexity}).

\subsection{Evaluation of Loss Functions}
The performance of two different loss functions was compared using the C-index across five cross-validation folds for BioFusionNet and MoCoV3 (Table \ref{tab:lossfunctionresult}). The results show that the mean performance improved for both methods when using the weighted Cox loss ($\mathcal{L}_\text{WCox}$) proposed in this paper,  compared to the traditional Cox loss ($\mathcal{L}_\text{Cox}$).

\begin{table}[!t]
\centering
\caption{Comparison of computational properties.}
\label{model_complexity}
\renewcommand{\arraystretch}{1.1} 
\resizebox{\columnwidth}{!}{%
\begin{tabular}{lrrr}
\toprule
Method               & Parameters & Size & FLOPS \\
\midrule
MultiSurv             & 1.79 M             & 6.99 MB     & 2.67 G \\
MultiDeepCox-SC             & 0.71 M             & 2.72 MB     & 1.18 G \\
HFBSurv               & 0.79 M           & 3.08 MB     & 2.65 G \\
PathomicFusion       & 102.28 M           & 397.12 MB   & 3.87 G \\
MCAT       &    2.06 M       & 8.06 MB   & 2.67 G \\
TransSurv       &    1.97 M       & 7.72 MB   & 2.67 G \\
BioFusionNet (Proposed) & 4.37 M            & 17.20 MB     & 12.11 G \\
\bottomrule
\end{tabular}}
\end{table}


\subsection{Univariate and Multivariate Hazard Analysis}
A comprehensive hazard analysis was conducted to evaluate the overall survival (OS) in the TCGA dataset of ER+ patients. Both univariate and multivariate analyses were performed (Table \ref{hazard-table1}). The analysis encompassed various parameters, including tumour grade, tumour size, age, lymph node (LN) status, subtype and the risk predictions made by BioFusionNet. In the multivariate analysis, positive LN status was associated with a hazard ratio (HR) of 1.87 (95\% CI: 1.32--2.64), demonstrating a significant effect on survival (p\,$<$\,0.005). Additionally, patients over the age of 55 had a HR of 1.77 (95\% CI: 1.07--2.91), also showing a significant impact on survival (p\,=\,0.03). However, no significant associations were found between tumour grade, size, or subtype and survival outcomes in this analysis. Notably, the BioFusionNet-predicted risk group (high vs.\ low) demonstrated a significant correlation with OS, with a HR of 2.91 (95\% CI: 1.80--4.68) (p\,$<$\,0.005). Univariate analysis indicated that tumour grade, size, age, and subtype were not statistically significant, whereas LN status (HR of 1.84, 95\% CI: 1.33--2.55, p\,$<$\,0.005) and BioFusionNet risk group (HR of 2.99, 95\% CI: 1.88--4.78, p\,$<$\,0.005) were significant predictors of survival. We note that the LN status had 51 missing values, which were imputed using a fixed value of 2. Kaplan-Meier survival analysis further supported the results, showing a significant difference in survival probabilities between the high- and low-risk groups as predicted by BioFusionNet (log-rank test p\,=\,6.45e-7) (Fig.~\ref{fig:kmplot}c).

\begin{table}[!t]
\smalltabcaption{Performance comparison of loss functions for cancer risk prediction using two different methods.}
\label{tab:lossfunctionresult}
\renewcommand{\arraystretch}{1.4} 
\resizebox{\columnwidth}{!}{%
\begin{tabular}{llcccccc}
\toprule
\multirow{2}{*}{Loss} & \multirow{2}{*}{Method} & \multicolumn{5}{c}{C-index (Fold)} & \multirow{2}{*}{Mean $\pm$ Std} \\ \cmidrule{3-7}
 & & 1 & 2 & 3 & 4 & 5 & \\
\midrule
\multirow{2}{*}{$\mathcal{L}_\text{Cox}$}
& BioFusionNet & 0.69  & 0.54  & 0.59  & 0.75 & 0.80 & $0.67\pm0.10$ \\
& MoCoV3 & 0.66  & 0.57 & 0.57 & 0.78 & 0.80 & $0.67\pm0.11$ \\
\midrule
\multirow{2}{*}{$\mathcal{L}_\text{WCox}$ (Proposed)}
& BioFusionNet & 0.78 & 0.71 & 0.72 & 0.81 & 0.82 & $0.77\pm0.05$ \\
& MoCoV3 & 0.70 & 0.66 & 0.66 & 0.77 & 0.72 & $0.70 \pm 0.04$ \\
\bottomrule
\end{tabular}%
}
\end{table}

\begin{table*}[!t]
\smalltabcaption{Univariate and multivariate analysis for overall survival (OS) in the TCGA dataset of ER+ patients.}
\label{hazard-table1}
\resizebox{\textwidth}{!}{%
\begin{tabular}{lllccccccc}
\toprule
\multirow{2}{*}{Parameter} & \multirow{2}{*}{Risk Group Cutoff} & \multirow{2}{*}{\#Patients/Group} & \multicolumn{3}{c}{Multivariate (n=249)} & & \multicolumn{3}{c}{Univariate (n=249)} \\
\cmidrule{4-6}\cmidrule{8-10}
& & & HR & 95\% CI & p & & HR & 95\% CI & p \\
\midrule
Tumour Grade & 3 vs.\ 1 \& 2 & 64 vs.\ 185 & 0.83 & 0.46--1.49 & 0.54 & & 1.13 & 0.67--1.91 & 0.65 \\
Tumour Size & \textgreater20 vs.\ $\leq$20 (mm) & 167 vs.\ 82 & 1.45 & 0.88--2.37 & 0.14 & & 1.37 & 0.86--2.19 & 0.19 \\
Age & \textgreater55 vs.\ $\leq$55 & 159 vs.\ 90 & 1.77 & 1.07--2.91 & 0.03 & & 1.47 & 0.91--2.36 & 0.11 \\
LN Status*
 & pos.\ vs.\ neg. & 110 vs.\ 88 & 1.87 & 1.32--2.64 & $<$0.005 & & 1.84 & 1.33--2.55 & $<$0.005\\
Subtype & lum B vs.\ A & 100 vs.\ 149 & 1.43 & 0.88--2.34 & 0.15 & & 1.38 & 0.88--2.18 & 0.16 \\
BioFusionNet & high vs.\ low & 132 vs.\ 117 & 2.91 & 1.80--4.68 & $<$0.005 & & 2.99 & 1.88--4.78 & $<$0.005 \\
\bottomrule
\multicolumn{10}{l}{\fontsize{8pt}{10pt}\selectfont*LN Status had 51 missing values which were imputed with a fixed value for both multivariate and univariate analysis.}  
\end{tabular}}
\vspace{-0.2\baselineskip}
\end{table*}

\subsection{Ablation Study}
We also evaluated the performance of various versions of BioFusionNet for ER+ breast cancer risk stratification. The results (Table \ref{tab:ablation}) show that the base model, BioFusionNet-B0 (MoCoV3, ViT Small) with $\mathcal{L}_{\text{WCox}}$, achieved the lowest C-index, and incorporating single cross-attention (SCA), dual cross-attention (DCA), or co-attention (CoA) yielded slight improvements, as did BioFusionNet-B1 (DINO33M, ViT Small) and BioFusionNet-B2 (DINO2M, ViT Small), both with DCA and $\mathcal{L}_{\text{WCox}}$. Combining BioFusionNet-B0, B1, and B2 with just $\mathcal{L}_{\text{WCox}}$ also resulted in slightly better performance than the base model, as did the inclusion of SCA and VAE. More substantial improvements of the combined model were obtained with the inclusion of DCA or CoA instead of SCA. The best performance was achieved by the combined model using VAE, CoA, DCA and late clinical data fusion ($C\text{-Fusion}_{\text{Late}}$) with $\mathcal{L}_{\text{WCox}}$, which clearly outperformed the same model using early clinical data fusion ($C\text{-Fusion}_{\text{Early}}$) or $\mathcal{L}_{\text{Cox}}$.

\begin{table}[!t]
\caption{Ablation study of BioFusionNet}
\label{tab:ablation}
\renewcommand{\arraystretch}{1.2} 
\resizebox{\columnwidth}{!}{%
\begin{tabular}{lc}
\toprule
Model & C-index \\
\midrule
BioFusionNet-B0 (MoCoV3, ViT Small) + $C\text{-Fusion}_{\text{Late}}$ + $\mathcal{L}_{\text{WCox}}$ & $0.65 \pm 0.05$ \\
BioFusionNet-B0 (MoCoV3, ViT Small) + SCA + $C\text{-Fusion}_{\text{Late}}$ + $\mathcal{L}_{\text{WCox}}$ & $0.69 \pm 0.04$ \\
BioFusionNet-B0 (MoCoV3, ViT Small) + DCA + $C\text{-Fusion}_{\text{Late}}$ + $\mathcal{L}_{\text{WCox}}$ & $0.70 \pm 0.03$ \\
BioFusionNet-B0 (MoCoV3, ViT Small) + CoA + $C\text{-Fusion}_{\text{Late}}$ + $\mathcal{L}_{\text{WCox}}$ & $0.70 \pm 0.04$ \\
BioFusionNet-B1 (DINO33M, ViT Small) + DCA + $C\text{-Fusion}_{\text{Late}}$ + $\mathcal{L}_{\text{WCox}}$ & $0.68 \pm 0.02$ \\
BioFusionNet-B2 (DINO2M, ViT Small) + DCA + $C\text{-Fusion}_{\text{Late}}$ + $\mathcal{L}_{\text{WCox}}$ & $0.67 \pm 0.03$ \\
BioFusionNet-Concat(B0+B1+B2) + $C\text{-Fusion}_{\text{Late}}$ + $\mathcal{L}_{\text{WCox}}$ & $0.67 \pm 0.04$ \\
BioFusionNet-Concat(B0+B1+B2) + SCA + $C\text{-Fusion}_{\text{Late}}$ + $\mathcal{L}_{\text{WCox}}$ & $0.69 \pm 0.03$ \\
BioFusionNet-Concat(B0+B1+B2) + VAE + SCA + $C\text{-Fusion}_{\text{Late}}$ + $\mathcal{L}_{\text{WCox}}$ & $0.68 \pm 0.04$ \\
BioFusionNet-Concat(B0+B1+B2) + VAE + DCA + $C\text{-Fusion}_{\text{Late}}$ + $\mathcal{L}_{\text{WCox}}$ & $0.75 \pm 0.04$ \\
BioFusionNet-Concat(B0+B1+B2) + VAE + CoA + $C\text{-Fusion}_{\text{Late}}$ + $\mathcal{L}_{\text{WCox}}$ & $0.70 \pm 0.03$ \\
BioFusionNet-Concat(B0+B1+B2) + VAE + CoA + DCA + $\text{Clinic-F}_{\text{Late}}$ + $\mathcal{L}_{\text{Cox}}$ & $0.67 \pm 0.10$ \\
BioFusionNet-Concat(B0+B1+B2) + VAE + CoA + DCA + $C\text{-Fusion}_{\text{Early}}$
+ $\mathcal{L}_{\text{WCox}}$ & $0.74 \pm 0.04$\\
BioFusionNet-Concat(B0+B1+B2) + VAE + CoA + DCA + $C\text{-Fusion}_{\text{Late}}$ + $\mathcal{L}_{\text{WCox}}$ & $0.77 \pm 0.03$ \\
\bottomrule
\end{tabular}}
\vspace{-0.5\baselineskip}
\end{table}

\subsection{Interpretability of BioFusionNet}
\begin{figure*}[!t]
\centering
\includegraphics[width=0.72\textwidth]{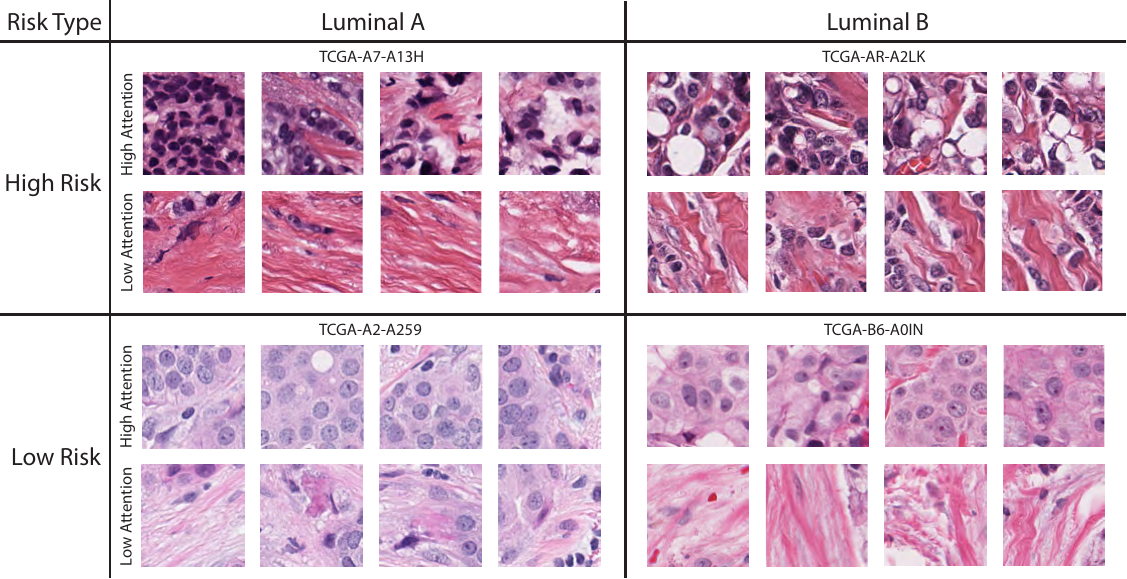}
\smallfigcaption{Visualisation of model-derived attention regions and associated risk types in Luminal A and Luminal B breast cancer patients. The figure presents raw histopathological image patches processed with BioFusionNet, which identifies areas of high and low attention, subsequently categorising patients into high and low risk.}
\label{fig:attentionpatch}
\end{figure*}

\begin{figure}[!t]
\centering
\includegraphics[width=0.5\textwidth]{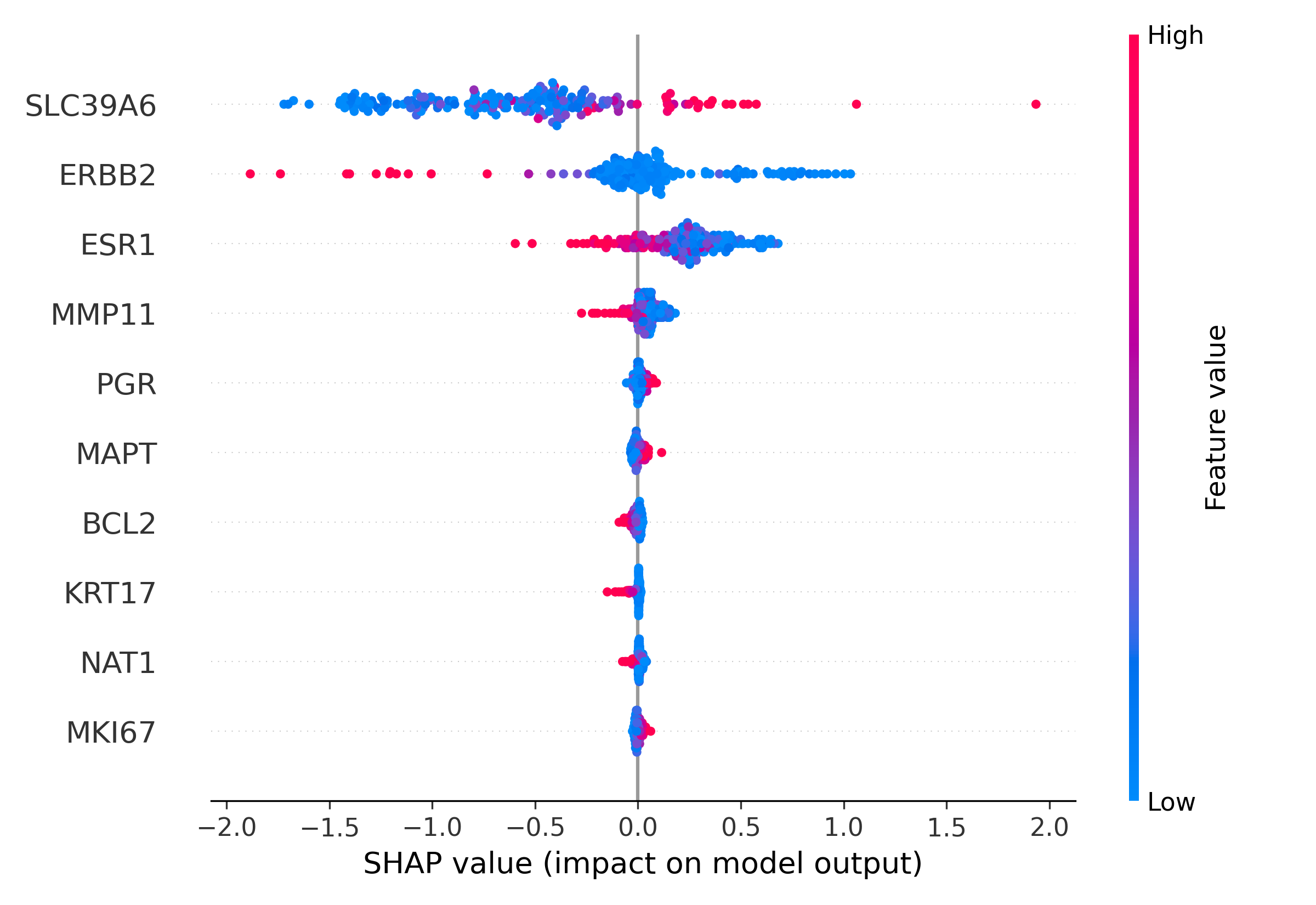}
\smallfigcaption{SHAP analysis of genetic features. The x-axis represents the SHAP value; colour intensity indicates gene expression level. The plot is sorted vertically by the features' overall importance.}
\label{fig:shapegene}
\end{figure}

\begin{figure}[!t]
\centering
\includegraphics[width=0.5\textwidth]{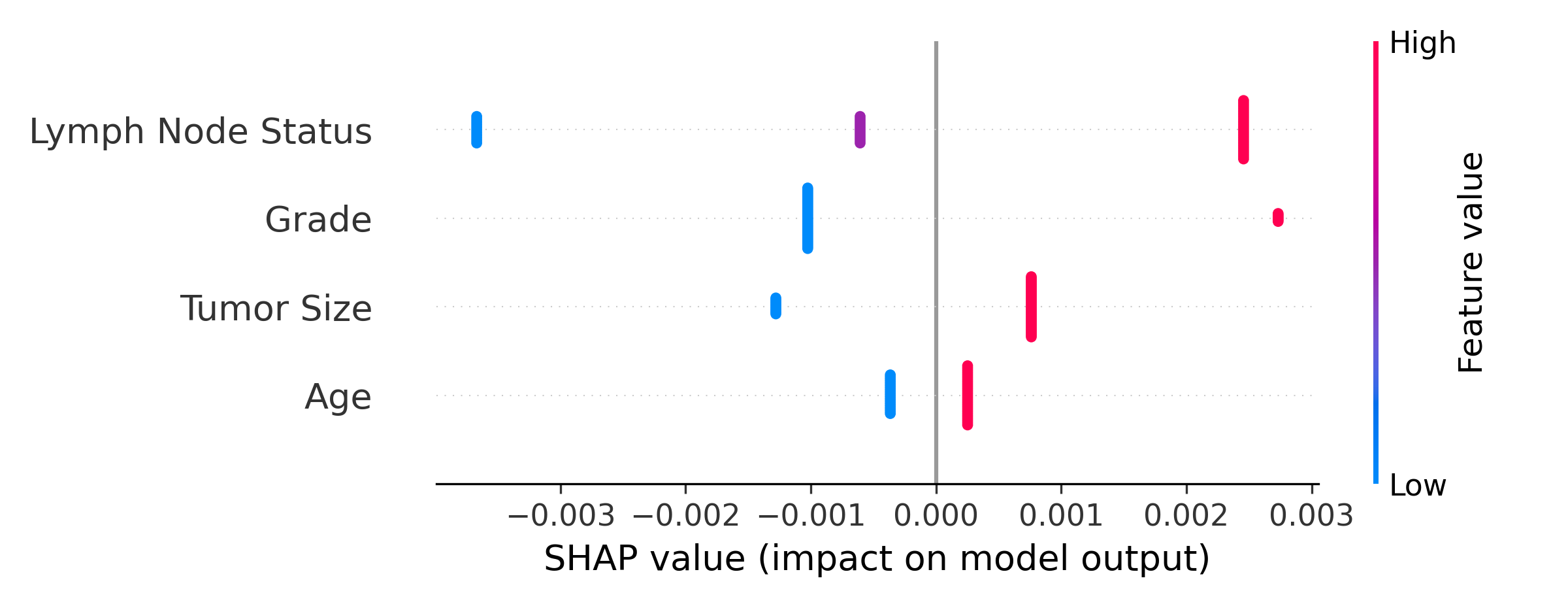}
\smallfigcaption{SHAP value distribution of clinical features. In this distribution, higher clinical values showing positive impact on the model, as indicated by its SHAP values.}
\label{fig:shapclinical}
\end{figure}

BioFusionNet utilises a self-attention mechanism to analyse histopathological image patches, identifying regions of high and low attention within both high-risk and low-risk patient profiles. Visual inspection of the results (Fig.~\ref{fig:attentionpatch}) reveals that regions with high attention contain distinct cellular patterns crucial for synthesising features from patch-level to patient-level, whereas areas of low attention typically exhibit less cellular atypia. This shows the model capacity to pinpoint clinically relevant features within tissue morphology. Additionally, SHAP analysis (Fig.~\ref{fig:shapegene}) reveals the influence of individual genes on the model predictions, ranked from high to low, providing interpretability of the risk assessment process. From this analysis, gene SLC39A6 (an estrogen regulated Zinc transporter protein with a role in epithelial to mesenchymal transition (EMT)) was identified as the most important predictor, with high expression levels producing high SHAP values, indicating positive impact on the model's cancer risk prediction. Other influential genes include ERBB2 (the gene for HER2), ESR1 (the gene for ER), with low expression levels producing high SHAP values therefore positive impact on the model. Moreover, the distribution of SHAP values for clinical features (Fig.~\ref{fig:shapclinical}) indicates that higher values of clinical parameters—such as positive LN status, higher tumour grade, increased tumour size, and postmenopausal age group—tend to have a positive impact on the model's output. In this context, a `positive impact' implies that the model associates these values with a higher likelihood of predicting patients at high risk.

\section{Discussion and Conclusion}
As demonstrated by the experimental results, the proposed BioFusionNet is highly effective for cancer risk prediction, showing superior performance compared to alternative approaches. Clearly, the multimodal fusion of imaging, genetic and clinical data allows the model to achieve substantially higher C-index scores compared to unimodal and dual-modal configurations, as well as compared to the traditional unimodal CoxPH and MLP. Furthermore, BioFusionNet outperforms existing multimodal fusion methods such as MultiSurv, HFBSurv, PathomicFusion, MCAT and TransSurv, and achieves the highest mean C-index ($0.77\pm0.05$) and AUC ($0.84\pm0.05$). Partly, the superior performance of BioFusionNet is due to the introduction of the proposed weighted Cox loss function instead of using the traditional Cox loss. Univariate and multivariate analyses showed the significant impact of age and BioFusionNet predictions on survival outcomes, while other clinical parameters such as tumour grade, size, lymph node status and subtype did not exhibit a significant correlation with survival outcomes. Kaplan-Meier analysis revealed a distinct separation in survival probabilities between the high-risk and low-risk groups identified by BioFusionNet. In addition, the results of the ablation experiment confirmed the importance of attention mechanisms in improving prediction accuracy, with the combined model configuration utilising VAE, CoA and DCA and the weighted Cox loss showing the highest performance.

BioFusionNet also presents a significant advancement in the interpretation of histopathological images, leveraging a self-attention mechanism to distinguish critical regions in patient profiles. A key contribution is the model's ability to align high-attention areas with distinct cellular patterns, crucial for transitioning from patch-level to patient-level analysis, thereby enhancing the diagnostic process. SHAP analysis amplifies this by clarifying the influence of specific genes and clinical features on the model's predictions. 
We observed that elevated SLC39A6 gene expression correlates with a high-risk prediction in ER+ breast cancers, where previous studies have shown conflicting findings, associating high SLC39A6 levels with good prognosis~\cite{Althobiti2021,Liu2020}, while others associated it with increased proliferation and lymph node involvement~\cite{Hogstrand2013,Gerold2016, Taylor2007}. Similarly, our model identified high ESR1 expression as indicative of low risk, aligning with literature that associates ESR1 positivity with enhanced responsiveness to endocrine therapy and, consequently, a better prognosis in ER+ breast cancer patients~\cite{Sappok2011}. In contrast, our analysis revealed an unexpected association between ERBB2 overexpression and a favorable prognosis in ER+ breast tumours, contrasting with the established view that ERBB2 overexpression indicates a poor prognosis~\cite{Xu2002, Xia2006}. As our study was specifically tailored to analyze ER+ samples, excluding the HER2-enriched subtype (known for its high ERBB2 expression and aggressiveness) likely influenced the findings. Moreover, the SHAP analysis for clinical factors (LN positivity, higher tumour grade and size, and postmenopausal age) significantly influences our model's ability to identify patients at increased risk, highlighting the critical role of these factors in breast cancer prognosis. Our analysis provides a transparent understanding of how each gene and clinical feature contributes to the model's predictions, providing actionable insights for clinical decision-making. While the risk assessment process mirrors current clinical practice, BioFusionNet streamlines the integration of all available data (patient features, tumour features and molecular features) to derive an automated single risk prediction score as a potential clinical oncology tool of the future.

While insightful, this study has certain limitations. We primarily opted for OS as the key outcome measure, instead of disease-free survival (DFS). This choice was made because DFS presented challenges such as a lower rate of events and a higher degree of data censorship, which could have limited the depth of the analysis. While OS is a feasible choice, it potentially overlooks critical insights into early-stage disease progression, typically highlighted by DFS. Moreover, the study's reliance on specific datasets such as TCGA for ER+ patients may affect the broad applicability of our findings. Another shortcoming of this study is the inherent limitations of the clinical data, which, during univariate analysis, identified tumour size, grade, and age as insignificant while only LN Status emerged as significant. Despite its limitations, the effectiveness of deep learning algorithms in analysing this clinical data arises from their ability to uncover complex patterns and interactions within dataset. Future research should therefore aim to validate these findings across a wider range of datasets to bolster the model's generalisability. Incorporating organ-level data, such as mammograms, could further enhance the predictive accuracy of our model. Additionally, extending the application of BioFusionNet to other cancer types and clinical scenarios could yield more comprehensive insights, making the research more universally relevant and applicable. Finally, a limitation of our model is that it is computationally demanding, primarily due to its extensive use of attention mechanisms. Whether the better performance justifies the higher computational cost depends on user needs and resources. Further research may provide ways to reduce the computational requirements of the model while retaining its high performance.

\section*{Data Availability}
TCGA image data and clinical data are publicly available at \url{https://portal.gdc.cancer.gov/}.

\section*{Code Availability}
Our work is fully reproducible and source code is publicly available on GitHub at \url{https://github.com/raktim-mondol/BioFusionNet}.

\section*{Acknowledgement}
This research was undertaken with the assistance of resources and services from the National Computational Infrastructure (NCI), which is supported by the Australian Government. Additionally, data preprocessing was performed using the computational cluster Katana, which is supported by Research Technology Services at UNSW Sydney.
\section*{Compliance with Ethical Standards}
This research study was conducted retrospectively using human subject data made available in open access by The Cancer Genome Atlas Breast Cancer (TCGA-BRCA) dataset, accessible through the National Cancer Institute's \href{https://portal.gdc.cancer.gov/projects/TCGA-BRCA}{Genomic Data Commons (GDC)} portal. Ethical approval was not required for this study, in accordance with the \href{https://www.cancer.gov/ccg/research/genome-sequencing/tcga/history/ethics-policies}{ethical policies} set forth by The Cancer Genome Atlas program.
\footnotesize
\bibliography{references}


\custombio{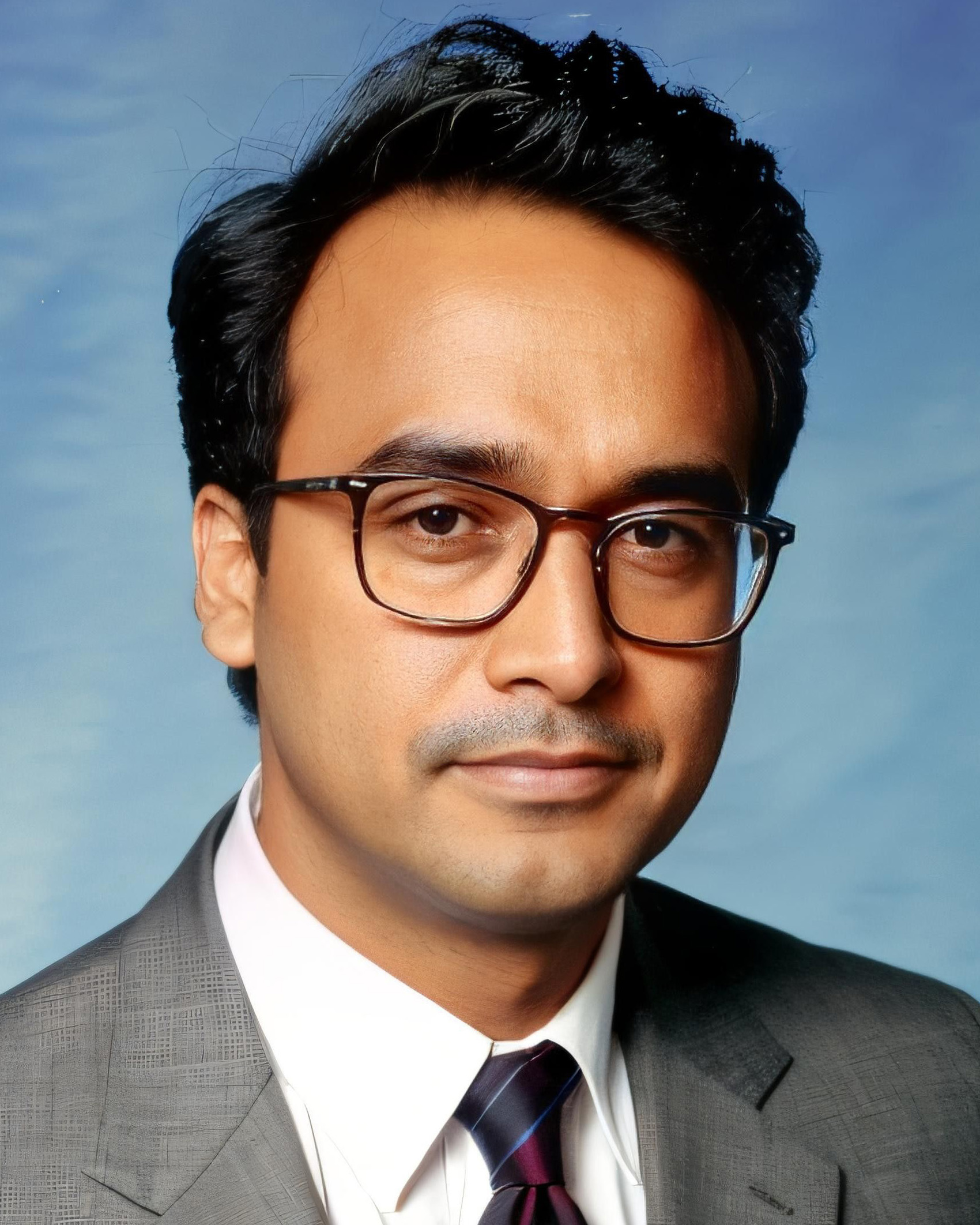}{\href{https://research.unsw.edu.au/people/mr-raktim-kumar-mondol}{Raktim Kumar Mondol}}{is a PhD candidate in Computer Science and Engineering, specializing in computer vision and bioinformatics. He completed his MEng in Engineering with High Distinction from RMIT University, Australia. Mondol's research interests include histopathological image analysis, clinical prognosis prediction, and enhancing clinical understanding through the interpretability of computational models.}

\custombio{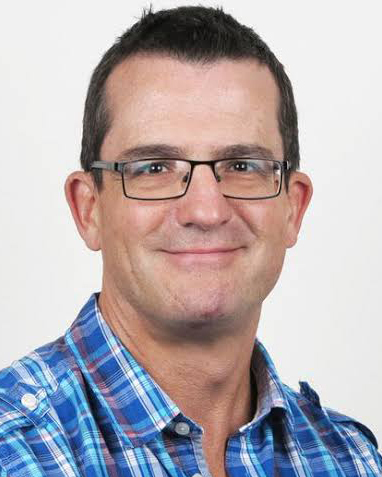}{\href{https://www.unsw.edu.au/staff/ewan-millar}{Ewan Millar}}{is a Senior Staff Specialist Histopathologist with NSW Health Pathology at St George Hospital Sydney with expertise in breast cancer pathology and translational research and a strong interest in AI and digital pathology applications.}

\custombio{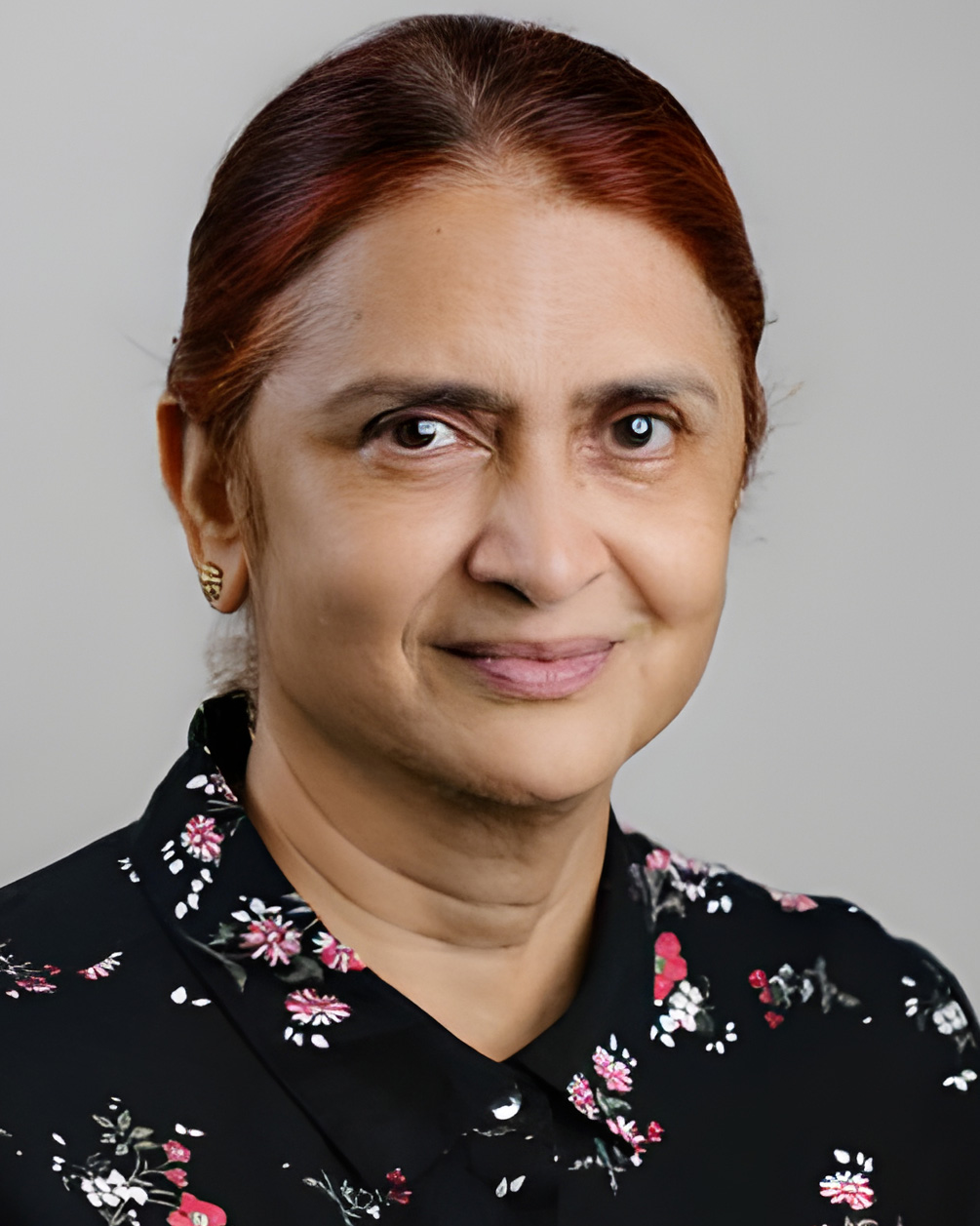}{\href{https://research.unsw.edu.au/people/professor-arcot-sowmya}{Arcot Sowmya}}{is Professor in the School of Computer Science and Engineering, UNSW. Her major research interest is in the area of Machine Learning for Computer Vision and includes learning object models, feature extraction, segmentation and recognition based on computer vision, machine learning and deep learning. In recent years, applications in the broader health area are a focus, including biomedical informatics and rapid diagnostics in the real world. All of these areas have been supported by competitive, industry and government funding.}

\custombio{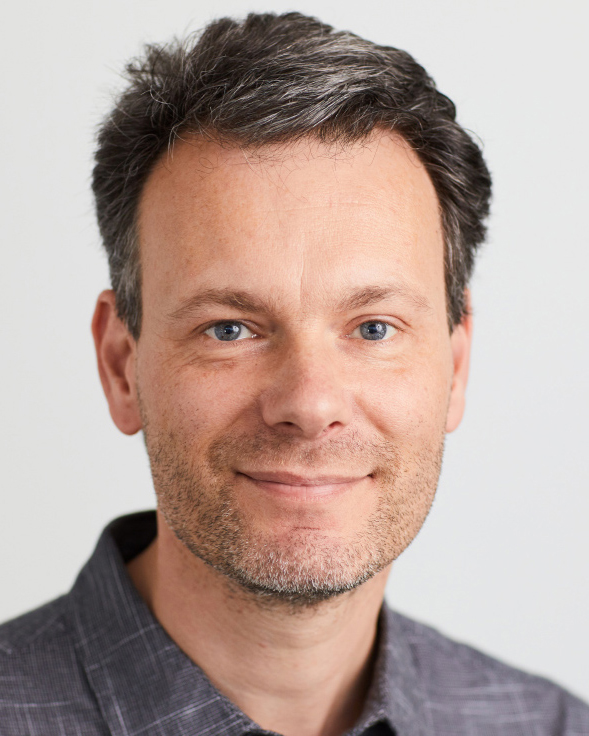}{\href{https://imagescience.org/meijering/}{Erik Meijering}}{(Fellow, IEEE), is a Professor of Biomedical Image Computing in the School of Computer Science and Engineering. His research focusses on the development of innovative computer vision and machine learning (in particular deep learning) methods for automated quantitative analysis of biomedical imaging data.}

\end{document}